\pdfoutput=1

\documentclass[11pt]{article}

\usepackage{acl}

\usepackage{times}
\usepackage{latexsym}
\usepackage{amsmath}
\usepackage{multirow}
\usepackage{bbm}
\usepackage{enumitem}
\usepackage{booktabs}
\usepackage{pifont}
\usepackage{array}
\usepackage{graphicx}
\usepackage{amsfonts}
\usepackage{hyperref}

\usepackage{subcaption}

\usepackage[T1]{fontenc}

\usepackage[utf8]{inputenc}

\usepackage{microtype}

\usepackage{inconsolata}

\usepackage{graphicx}

\title{Towards Faithful Natural Language Explanations: A Study Using Activation Patching in Large Language Models}

\author{Wei Jie Yeo\textsuperscript{1}, Ranjan Satapathy\textsuperscript{2}, Erik Cambria\textsuperscript{1}\\
\textsuperscript{1}Nanyang Technological University \\
\textsuperscript{2}Institute of High Performance Computing (IHPC),\\ Agency for Science, Technology and Research (A$\textasteriskcentered$ STAR), Singapore\\
\texttt{yeow0082@e.ntu.edu.sg} \\}

\begin{document}
\maketitle
\begin{abstract}
Large Language Models (LLMs) are capable of generating persuasive Natural Language Explanations (NLEs) to justify their answers. However, the faithfulness of these explanations should not be readily trusted at face value. Recent studies have proposed various methods to measure the faithfulness of NLEs, typically by inserting perturbations at the explanation or feature level. We argue that these approaches are neither comprehensive nor correctly designed according to the established definition of faithfulness. Moreover, we highlight the risks of grounding faithfulness findings on out-of-distribution samples. In this work, we leverage a causal mediation technique called activation patching, to measure the faithfulness of an explanation towards supporting the explained answer. Our proposed metric, \textit{Causal Faithfulness} quantifies the consistency of causal attributions between explanations and the corresponding model outputs as the indicator of faithfulness. We experimented across models varying from 2B to 27B parameters and found that models that underwent alignment-tuning tend to produce more faithful and plausible explanations. We find that \textit{Causal Faithfulness} is a promising improvement over existing faithfulness tests by taking into account the model’s internal computations and avoiding out-of-distribution concerns that could otherwise undermine the validity of faithfulness assessments. We release the code in \url{https://github.com/wj210/Causal-Faithfulness}
\end{abstract}

\section{Introduction}
\label{lab:intro}
The advent of Large Language Models (LLMs) has revolutionized the field of Natural Language Processing (NLP), enabling machines to perform considerably well across a wide array of tasks ranging from commonsense reasoning~\citep{achiam2023gpt} to producing high-quality training data~\citep{jie2024selftraining}. Moreover, LLMs can generate highly plausible Natural Language Explanations (NLE) to support their answer, such as adopting the Chain-of-Thought (CoT) prompting method~\citep{wei2022chain}. Despite their apparent sophistication, the trustworthiness of these explanations is not guaranteed and their underlying faithfulness~\citep{jie2024interpretable} should be carefully scrutinized. 

However, it is unclear amongst the existing approaches~\citep{atanasova2023faithfulness,lanham2023measuring,wiegreffe2020measuring,parcalabescu2023measuring}, which is best suited for measuring faithfulness. A commonly referenced definition of faithfulness follows as such: "\textit{The faithfulness of an interpretation is a measure of how well it accurately reflects the underlying reasoning process of the model}"~\citep{jacovi2020towards}. Following this definition, we can be certain that a truly faithful explanation leads to goals such as instilling trust~
\citep{cambria2023seven,yeo2023comprehensive} in the receiving audience or enabling actionable insights given the information revealed by the explanation.

A recent work, \citet{parcalabescu2023measuring} argues that existing faithfulness tests are not designed to measure the faithfulness under the referenced definition, but rather the ability of models to return consistent answers under various perturbed settings. Albeit these tests are still considered valid, the findings of these tests may be unreliable and inconsistent across different benchmarks. In this work, we study the internal computations of the model to assess the level of faithfulness of an explanation in supporting the explained answer. We borrow insights from works that perform Activation Patching (AP), also known as causal tracing on the LLM's internal states~\citep{meng2022locating,vig2020causal,zhang2023towards}, and show how the derived causal effects can be used to measure the faithfulness of a model's NLE.

Our metric, Causal Faithfulness (CaF) compares the distribution over the causal effects behind the answer and NLE in a similar fashion as ~\citet{parcalabescu2023measuring}. However, we explore at a deeper level, by also examining consistency at the layer positions and perform AP instead of SHAP~\citep{lundberg2017unified} which we later show is prone to Out-Of-Distribution (OOD) issues. We find that CaF is by design, a closer test of establishing the true faithfulness of NLEs.
Overall, the main contributions of this work can be summarized as follows:
\begin{enumerate}
    \item We first introduce \textit{Causal Faithfulness}, which utilizes activation patching in place of SHAP, to investigate faithfulness at three levels: token and layer, token-only and layer-only. 
    \item We investigate the relationship between the plausibility and faithfulness of NLEs to determine if a plausible NLE can be equally regarded as faithful.
    \item We analyze our approach extensively against existing faithfulness tests on $6$ open-sourced LLMs of varying sizes and across $3$ benchmarks.
\end{enumerate}

\section{Related Works}
\subsection{Faithfulness of NLP models}
NLE is an abstraction form of explanation that consists of a sequence of free-text tokens, aimed at explaining the internal reasoning process of the model. On the other hand, extractive approaches~\citep{deyoung2019eraser,lei2016rationalizing,jie2024plausible} highlight rationale tokens in the input to serve as the explanation. Though extractive approaches may appear to be faithful, ~\citet{jacovi2021aligning} cautions against such beliefs as such an approach requires the explanation to be produced before the answer, thereby questioning its soundness. Existing works on faithfulness introduce different forms of perturbations such as paraphrasing or corrupting the CoT explanation~\citep{jie2024interpretable,lanham2023measuring}, or checking for consistency in counterfactuals~\citep{atanasova2023faithfulness}. ~\citet{wiegreffe2020measuring} corrupts the input and regards the similarity between the deterioration of the answer and explanation as a proxy for faithfulness.~\citet{turpin2024language} showed that adding biasing features can significantly influence the CoT leading to incorrect answers.

\citet{parcalabescu2023measuring} expanded on SHAP~\citep{lundberg2017unified} by introducing CC-SHAP, which measures the convergence between the attribution vectors of the answer and explanation as a measure of faithfulness. The authors argue that CC-SHAP more closely aligns with the definition of faithfulness as '\textit{how well the explanation reflects the reasoning process}.' Our approach, CaF, is similar to CC-SHAP but operates at a deeper level by also considering convergence between the layers while avoiding OOD samples which may invalidate any findings.

\subsection{Causality in NLP}
Causal mediation methods aim to quantify the sensitivity of intermediate variables within a causal graph, following an intervention. The sensitivity is termed as causal effects and can be further decomposed into direct and indirect effects~\citep{pearl2022direct}. Works leveraging causality insights include~\citet{gat2023faithful}, which assesses the faithfulness of NLP models based on their ability to identify features with significant causal effects. The authors argue that inputs with high causal influence are more likely to be successfully transformed into counterfactuals. ~\citet{vig2020causal} detects gender bias by measuring the causal effects of a stereotypical input given an intervention.~\citet{paul2024making} investigates the direct and indirect effects in CoT explanations generated by LLMs of different sizes and concludes that causal effects of CoT are task-dependent, complimenting our findings.~\citet{meng2022locating} perform causal mediation on the hidden states of a transformer~\citep{vaswani2017attention} and discover that certain states are responsible for recollecting factual knowledge. The authors then showed that intervening in the states with the highest causal effect is helpful in knowledge editing. This approach inspired our work, where we perform similar interventions to quantify the causal effects of each LLM's internal states during test-time as a means of measuring faithfulness.

\section{Preliminary}
\begin{figure*}[ht]
    \centering 
    \includegraphics[width=\textwidth]{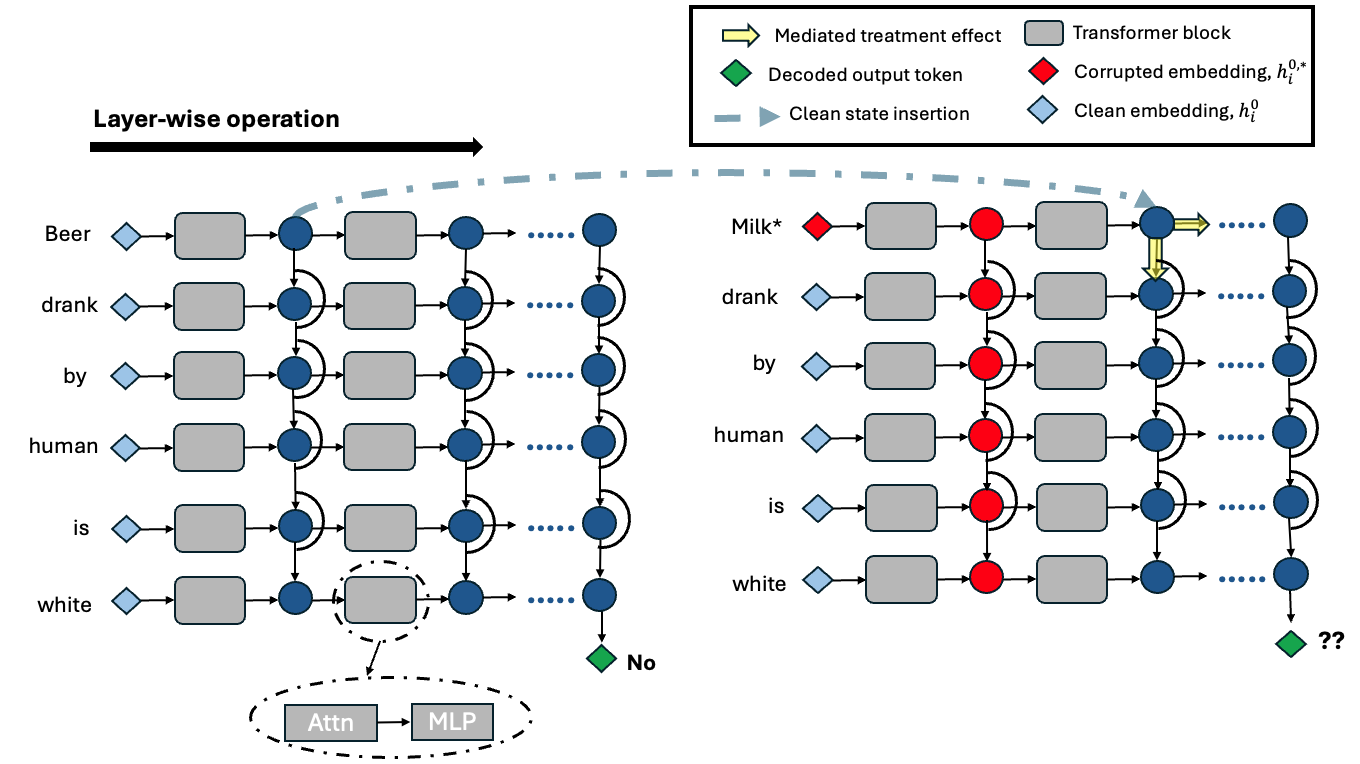}
    \caption{Activation Patching: Given two runs, a clean run under normal conditions [left] and a corrupted run where tokens in the input are replaced such that it leads to a counterfactual scenario. [right]. AP identifies the causal effects of the hidden state at the specified token and layer position through the changes in output after inserting the activations from the clean run. The indirect effect is thus measured via the mediated effects of the intervention.~\citep{meng2022locating}.} 
    \label{fig:causal tracing}
\end{figure*}
We focus on decoder-only transformers LLMs, parameterized as $f_{\theta}$ that is $L$-layers deep, each consisting of both an Attention (attn) and Multi-Layer Perceptron (MLP) module. $f_{\theta}$ models the discrete probability distribution of the next token over a predefined vocabulary set $V$ in an autoregressive manner, $f_{\theta}(p(x_{t+1})|x_t,\ldots,x_1$), where $p(x_{t+1}) \in \mathbb{R}^{|V|}$. The output token, $\hat{y} = x_{t+1}$ is then sampled from the probability distribution. For each input question, $f_{\theta}$ generates both an answer, $\hat{y}_a$ and a post-hoc explanation, $\hat{y}_e$. We are interested in assessing if $\hat{y}_e$ is a faithful representation of the internal reasoning responsible for $\hat{y}_a$. We denote the intermediate hidden representation of the token $i$ at the $l$ layer as $h_i^l \in \mathbb{R}^{|K|}$, $K$ referring to the hidden size. The text tokens are encoded into embeddings with positional information represented as $h_i^0 = f_{\theta,embed}(x_i) + pos(x_i)$. The internal computations of $h_i^l$ consist of a self-attention, $a_i$ followed by an MLP operation, $m_i$ is described as:
\begin{align}
\label{eq:llm}
h^{l}_i &= h^{l-1}_i + a^{l}_i + m^{l}_i \notag \\
a^{l}_i &= \text{attn}^{\hspace{1pt}l}\left( h^{l-1}_1, h^{l-1}_2, \ldots, h^{l-1}_i \right) \notag \\
m^{l}_i &=  \text{MLP}^{\hspace{1pt}l}(a^{l}_i + h^{l-1}_i)
\end{align}
\subsection{Activation Patching}
\citet{meng2022locating} applied AP to identify specific hidden states essential for storing factual knowledge, which later proved useful for fact-editing. AP primarily requires three forward passes: a clean, $p(\hat{y})$, corrupted, $p_*(\hat{y})$ and patched, $p_{*,clean\hspace{2pt} h_i^l}(\hat{y})$ run at the specified \textit{i} and \textit{l} hidden state positions. \citet{meng2022locating} corrupts the embedding vectors of the subject tokens in the input by adding Gaussian noise (GN), $h_i^{0,*} = h_i^0 + \epsilon$, sampled from $\mathcal{N}(0,3 \sigma)$, $\sigma$ given as the standard deviation between the set of input embeddings over the subject tokens, with indices belong to a span, $S$ of varying length. The patched run then intervenes at the \textit{i} token and \textit{l} layer position by patching in the hidden states from the clean run as seen in Figure~\ref{fig:causal tracing}, before continuing until the final output. This is repeated across all layers and input tokens such that we end up with a matrix, $C \in \mathbb{R}^{T \times L}$, and each element represents the causal effects of the corresponding hidden state towards each output generation, $\hat{y}_a$ or $\hat{y}_e$. Note that patching states at token positions $i$ such that $i < j$ for all $j \in S$ is pointless given the nature of causal attention used in decoder models. Thus $T$ represents the length of the original input after excluding the token positions preceding $S$.
\begin{figure*}[ht]
    \centering
    \begin{subfigure}[b]{0.4\textwidth}
        \centering
        \includegraphics[width=\textwidth]{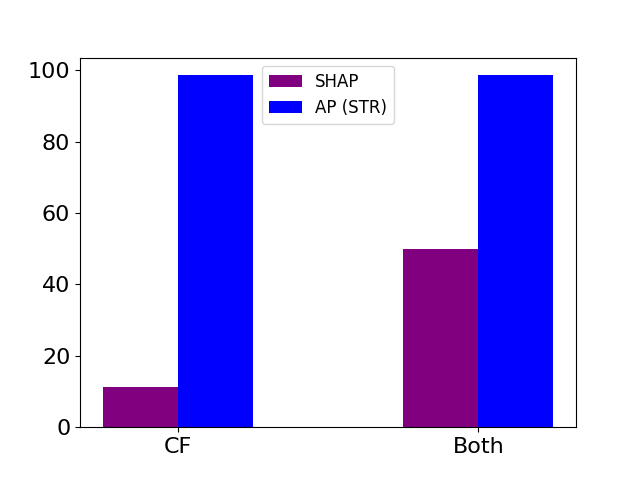}
        \caption{SHAP vs AP (STR)}
        \label{fig:ood_shap}
    \end{subfigure}
    \begin{subfigure}[b]{0.4\textwidth}
        \centering
        \includegraphics[width=\textwidth]{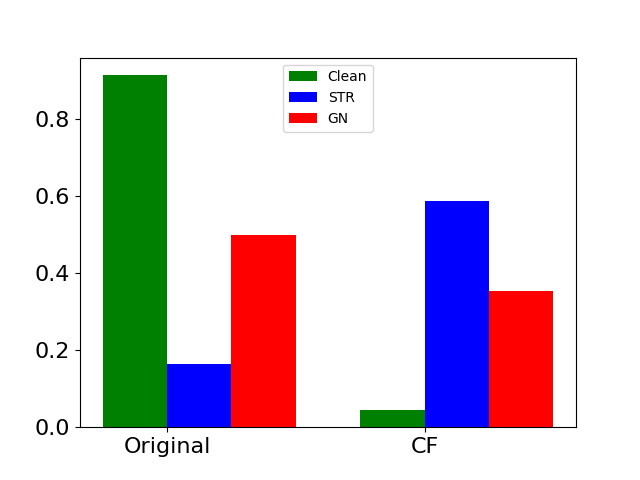}
        \caption{STR vs GN}
        \label{fig:ood_gn}
    \end{subfigure}
    \caption{[Left] Counts of instances where the modified features are assigned higher importance. [Right] Probability scores of original and counterfactual answers in the clean and corrupted (STR/GN) runs.}
    \label{fig:ood}
\end{figure*}

Based on ~\citet{pearl2022direct}, we can retrieve various causal effects from these three runs. The total effect due to the corruption is given as $p(\hat{y}) - p_*(\hat{y})$. We are interested in the \textbf{indirect effect} of the target hidden state, given as $p_{*,clean\hspace{2pt} h_i^l}(\hat{y}) - p_*(\hat{y})$, where the mediator (hidden state) is intervened while the independent variable (subject token) is held at the no-treatment state. This mediated effect can be observed in Equation~\ref{eq:llm}, layer-wise via the MLP and position-wise in the self-attention module. However, adding GN is prone to inducing OOD inputs~\citep{zhang2023towards} that may invalidate the attribution readings. This is particularly problematic as $C$ should ideally represent the importance of the target state in generating the dependent variable, rather than relieving the model from the distribution shift. We perform Symmetric Token Replacement (STR) instead, by replacing a subset of tokens in the input such that it becomes a counterfactual instance. We find that this relieves the OOD issue since this process is controlled as $y$ can be converted into counterfactual alternatives, $y_C \neq y$, while adding GN is uncontrollable. We discuss more in ~\ref{sec:ood}.
\section{Methodology}
In this section, we discuss why activation patching is a suitable method for measuring the faithfulness of NLE. While the causal values from AP itself suffices as an explanation, it may appear less interpretable for human users, particularly when multiple hidden states exhibit significant causal effects. This is especially true when the output space, $T$, is large, as the resulting explanation can highlight numerous factors contributing to the model's decision. Such explanations, though accurate, may seem less intuitive compared to natural language explanations, which are generally easier for users to understand and process. On the other hand, NLEs should not be trusted easily as their faithfulness is questionable. With the abundance of existing faithfulness tests, this poses another question of which is best suited to assess the faithfulness of NLE? 

\begin{figure*}[ht]
    \centering 
    \includegraphics[width=0.9\textwidth]{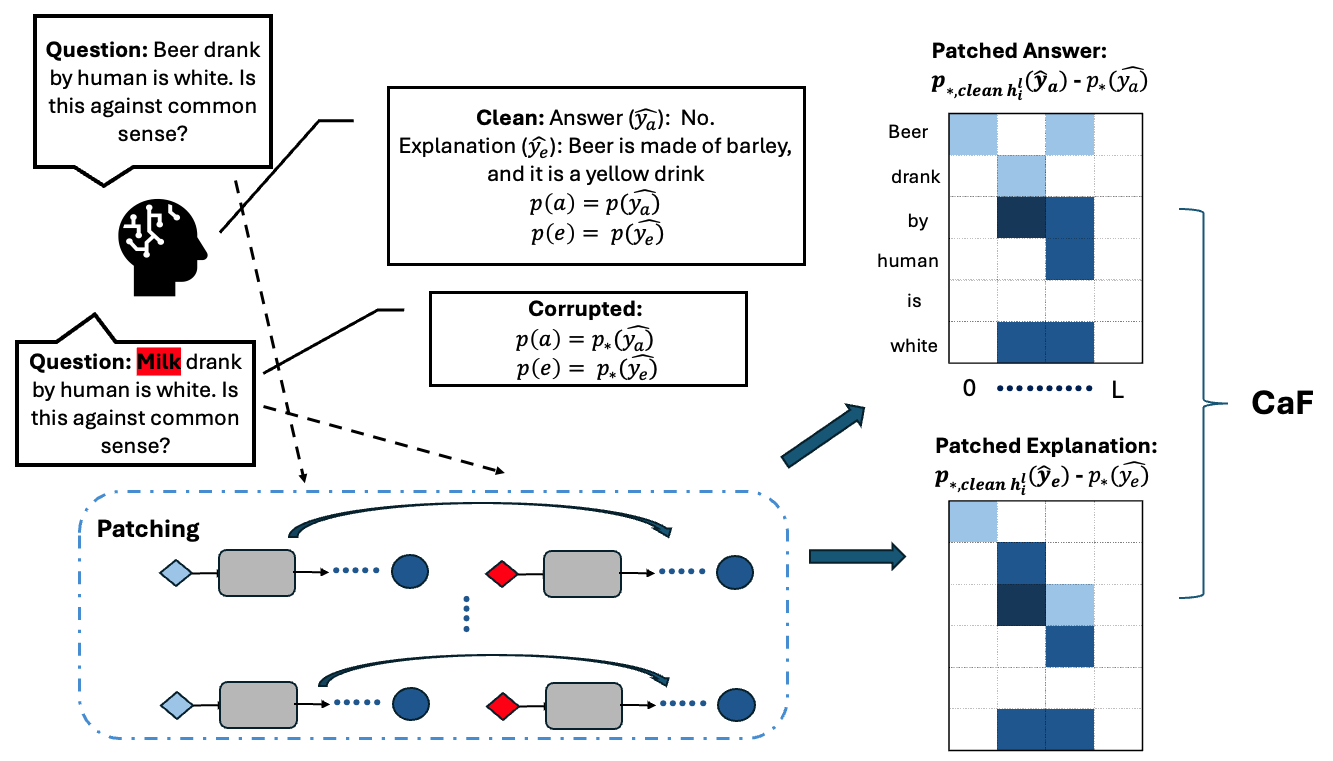}
    \caption{The model's probability scores from both clean corrupted runs are recorded and deducted from the patched scores over each token and layer. All activations from the clean run are hooked and subsequently patched in at the target location before continuing the run. AP is implemented for both outputs: answer and explanation, resulting in the final causal matrix $C$, before measuring CaF.} 
    \label{fig:caf}
\end{figure*}
~\citet{parcalabescu2023measuring} argues that most of these tests are designed to evaluate self-consistency rather than the referenced definition: \textit{"aligns with the underlying reasoning process behind the model's answer"}. Albeit asserting complete faithfulness is impossible as it would require reverse-engineering the internal mechanism of several billion parameters. It is also crucial to avoid erroneous measurements that may mislead the audience into attributing incorrect causes, which could have severe consequences in critical applications~\citep{agarwal2024faithfulness}. This phenomenon, referred to as \textit{social misalignment}~\citep{jacovi2021aligning,hase2021out}, occurs when the understanding of an explanation is distorted. Such misalignment can arise from various factors, including the evaluation of models on  OOD samples. ~\citet{parcalabescu2023measuring} introduces CC-SHAP by extending upon SHAP~\citep{lundberg2017unified} to measure the consistency between the Shapley values of $\hat{y_a}$ and $\hat{y_e}$. The authors regard this form of measurement as a closer step towards assessing faithfulness under the commonly referenced definition. 

\subsection{Implications of Out-of-Distribution samples}
\label{sec:ood}
In this section, we argue that using SHAP as an attribution technique to assess faithfulness has notable limitations. Similar to GN, we show that SHAP is susceptible to OOD samples due to its nature of integrating across all possible permutation pairs, this can lead to feature combinations likely unseen by the model. We investigate SHAP's behavior in attributing scores to counterfactual features that influence changes in model predictions.

We compare the consistency between SHAP and AP on the ComVE~\citep{wang2020semeval} task with Gemma2-2B-chat~\citep{team2024gemma}, where the model is required to identify the illogical statement from two given statements. We split the task into two scenarios: the original, where the illogical statement is presented, and the counterfactual (CF), where the logical statement is given. Both statements, each of length $T$, differ by a subset of tokens with positions in $S \subseteq T$. We count how often each attribution method assigns a higher average score to $S$ over the complement, $S^N = T \setminus S$. The analysis covers both the original and CF scenarios, where the model is instead given a logical statement. We only consider instances where the model is successful in both scenarios. Figure~\ref{fig:ood_shap} shows the averaged count of $S > S^N$. SHAP assigns higher importance to $S$ in only $12\%$ of the CF scenarios and $50\%$ of both when the model assigns higher importance in $S > S^N$ for the original as well. In contrast, AP assigns higher importance to $S$ in nearly $95\%$ of cases across both scenarios. We note that modifications to the feature space do not always result in the highest attributions in the event of a prediction shift. However, when the original input regards $S$ as important, failing to observe this when a prediction is altered, indicates inconsistency and raises concerns about the fidelity of the approach. This inconsistency presents a challenge when SHAP is used to assess faithfulness.

For GN, we notice that this form of corruption can lead to inputs that deteriorate model's understanding of the task rather than removing important information. In Figure~\ref{fig:ood_gn}, adding GN causes the model to assign similar probability scores between the two answers, while STR correctly swaps the scores from illogical (Original) to logical (CF). Since ComVE only has two possible outcomes, assigning an equal score indicates a lack of confidence in either outcome. We note that STR allows for greater control over the removal or negation of relevant information, thus enabling in-distribution corruption, as opposed to GN which may break the model's ability to understand the task correctly. This is analogous to the findings in ~\citet{zhang2023towards}, where GN is observed to disrupt the model's internal mechanism, which is not the intended objective of AP. 

\subsection{Causal Faithfulness}
We introduce Causal Faithfulness, by measuring the divergence between the causal matrix of the answer and explanation, $C_a$ and $C_e$ respectively. Each matrix is generated via activation patching over layers of length $L$ and feature set of size $T$ towards $\hat{y_a}$ and $\hat{y_e}$, see Figure~\ref{fig:caf}. For outputs with more than one token, we average over the output sequence, $M$ as given:
\begin{equation}
    C_i^{\hspace{1pt}l} = \frac{1}{M} \sum_{j=1}^{M} C_{i,j}^{\hspace{1pt}l}
\end{equation}
 We similarly employ cosine distance ($CD$) as our divergence metric, prioritizing the distribution of attribution over magnitude, since $C_a$ and $C_e$ may have significantly different ranges.
\begin{equation}
\label{eq:2}
    \text{CaF} = 1 - CD(C_e,C_a)
\end{equation}
Equation~\ref{eq:2} produces a continuous score, similar to CC-SHAP~\cite{parcalabescu2023measuring}, but also considers the layer-wise effects. We argue that assessing faithfulness according to the referenced definition, requires making judgments beyond just at the feature level. For instance, on reasoning tasks, one would expect the model to reason internally before generating $\hat{y_a}$, where the internal reasoning process is revealed via $\hat{y_e}$. Input-level attribution addresses the question of \textit{"which features are important"} but not \textit{"how does the model process each feature"}. While the notion of sufficient granularity for faithfulness is yet to be rigorously defined, layer-wise alignment can be particularly advantageous, such as performing corrective actions on models with undesirable bias in certain layers, potentially observed via the NLEs.
On the contrary, CaF does not assume that faithfulness is guaranteed purely at the feature level but rather states that true faithfulness should also consider the model's internal computations. 

\section{Experiments}
\subsection{Dataset and Model}
We implement the faithfulness test on 3 different benchmarks: CoS-E~\citep{rajani2019explain}(Commonsense reasoning), e-snli~\citep{camburu2018snli} (Natural Language Inference) and ComVE~\citep{wang2020semeval}. CosE and e-snli provides annotated rationales tokens, which we used to create counterfactuals. In ComVE, we swap the logical statement in to act as the counterfactual. We discuss more in ~\ref{sec:implementations}. 

We evaluate the Gemma-2 suite of LLMs~\citep{team2024gemma}, of three different sizes: 2B, 9B and 27B, including both pre-trained and instruct-tuned versions, totalling to 6 models. We use 3-shot prompting for pre-trained models and evaluate chat models under zero-shot settings. All explanations are generated post-hoc, conditioned on the model's answer to the question. We use the same template for each task for standardization. We assess a total of 100 samples for each task and average the scores across three seeds for each faithfulness metric.

\begin{table*}[ht]
\centering
\begin{tabular}{cc|ccccccc}
\toprule
 \multicolumn{2}{c|}{Model/Test}& Acc & CFF & CC-SHAP & CaF & CaF(M) & CaF(T) & CaF(L) \\
\midrule
\multirow{4}{*}{\rotatebox{90}{\textbf{CoS-E}}} & 2B & 42 & 14& 62 & 13 & 13 & 20 & 10 \\
& 2B-chat & 58 & 57 & 74 & 25 & 26 & 33 & 20\\
& 9B & 58& 20 & 35& 18 & 17 & 23 & 9\\
& 9B-chat & 76& 57& 89 & \textbf{39}& 37 & 49 & 35\\
& 27B & 68& 20& 40 & 34& 33 & 43 & 19\\
& 27B-chat & 76& 59& 15 & 37& 36 & 45 & 33\\
\midrule
\multirow{4}{*}{\rotatebox{90}{\textbf{e-snli}}} & 2B & 59 & 25& 65 & 8 & 7 & 14 & 5 \\
& 2B-chat & 64 & 56 & 64 & 11 & 12 & 12 & 8\\
& 9B & 82& 14& 37& 26& 25 & 30 & 21\\
& 9B-chat & 90& 64& 91 & \textbf{41}& 41 & 50 & 37\\
& 27B & 91& 13& 62 & 34& 36 & 42 & 16\\
& 27B-chat & 90& 63& 54 & 39& 42 & 40 & 27\\
\midrule
\multirow{4}{*}{\rotatebox{90}{\textbf{ComVE}}} &  2B & 52 & 17 & 82 & -6 & -9 & -6 & 5 \\
& 2B-chat & 69 & 28 & 98 & 44 & 44 & 54 & 43 \\
& 9B & 68& 23& 47 & 24& 24 & 33 & 17 \\
& 9B-chat & 91& 78& 91 & 40 & 37 & 53 & 35\\
& 27B & 86& 24& 82 & 39& 36 & 51 & 23\\
& 27B-chat & 95& 81& 76 & \textbf{51}& 47 & 61 & 46\\
\bottomrule
\end{tabular}
\caption{Faithfulness test scores across different metrics. \textbf{M} refers to patching across multiple layers, \textbf{T} is the aggregated effects across layers within the same token position and \textbf{L} across tokens within the layer. Instruct-tuned models on average tend to fare better as compared to the corresponding pre-trained counterparts. CaF and CC-SHAP is bounded between -1 and 1. Acc refers to the accuracy scores according to the task. All values are multiplied by 100.}
\label{tab:result_faithful}
\end{table*}

\subsection{Faithfulness Tests}
We also include other baseline faithfulness metrics designed for post-hoc explanations: CC-SHAP~\citep{parcalabescu2023measuring} and Counterfactual Faithfulness (CFF)~\citep{atanasova2023faithfulness}. CC-SHAP is similar to CaF but replaces AP with SHAP and measures the cosine similarity between the attribution vectors at the feature level. CFF introduces an adjective or adverb token at random positions in the input to influence the model into altering its original prediction, within a specified budget. If the model maintains its original prediction, it is considered faithful. CFF is primarily employed to identify instances of unfaithfulness, where the model's prediction changes, but the counterfactual explanation omits the inserted token. A noted limitation of CFF is that it focuses solely on syntactic checks, overlooking the importance of maintaining semantic consistency.

Additionally, we explore the impact of patching across multiple layers, rather than a single layer, which allows for a greater recovery.~\citep{meng2022locating,zhang2023towards}. We use a  window, $w$ of $10$, patching the layers between $[l-\frac{w}{2}$, $l+\frac{w}{2}]$ to represent the effects at layer $l$. We also study changes to CaF when aggregating effects across the layer and token position. We denote CaF(T) as the divergence between token-level causal vectors where the layers within each token position are aggregated and vice versa at the layer-level, CaF(L).

The focus of our experiment is to address the following research questions:
\begin{itemize}
    \item How do the NLEs from different LLMs fare under various faithfulness metrics
    \item Is there a correlation between a model's capability and the faithfulness of the explanations produced?
    \item Does a plausible explanation entail a higher degree of faithfulness?
\end{itemize}


\subsection{Findings}
\label{sec:faithfulness_tests}
\textbf{Chat models are more faithful}: Table~\ref{tab:result_faithful} indicates that instruct-tuned models tend to produce more faithful explanations, as indicated by CaF. This highlights the additional benefits of alignment tuning on faithfulness besides enhancing the plausibility of explanations, which we discuss further in ~\ref{sec:plaus}. Since pre-trained models are not fine-tuned for specific task structures, they may lack an internal understanding of the task, such as when required to explain their own decisions. This can lead to a discrepancy in the distribution of causal values between the generated answer, $\hat{y}_a$, and the explanation, $\hat{y}_e$, resulting in lower consistency between the two. Patching across multiple layers do not have significant differences, though we find that models exhibit a higher degree of consistency when the layer effects are aggregated as in CaF(T) and less so at the layer level, CaF(L). We also observe a positive relationship between task performance and faithfulness, when comparing within the training category (ie pre-trained or instruct-tuned), which applies similarly to model scaling as well.\\\\
\textbf{CaF vs other tests}: As mentioned by ~\citet{parcalabescu2023measuring}, a primary concern is the high level of disagreement between existing faithfulness tests. Besides that, in ~\ref{sec:ood}, we discuss the risks involved with encountering OOD samples depending on the form of attribution method used, that affects metrics such as CC-SHAP. For CFF, a drawback as highlighted by~\cite{siegel2024probabilities,parcalabescu2023measuring}, is that the scores can be manipulated by models trained to produce excessively verbose responses, where the input is always replicated in the explanation. Given the design of the test, this would almost certainly result in achieving perfect faithfulness. Another problematic concern is that the counterfactual edits may not always result in a modified prediction, which would default to an indication of faithfulness. However, we argue that this has nothing to do with the faithfulness of NLEs, but rather points towards the robustness of the model against adversarial inserts. In our experiments, we note that this occurrence is rather significant in instruct-tuned models, ranging from $10\%$ to $70\%$ of the dataset.\\\\

\begin{figure*}[ht]
    \centering
    \includegraphics[width=0.8\textwidth]{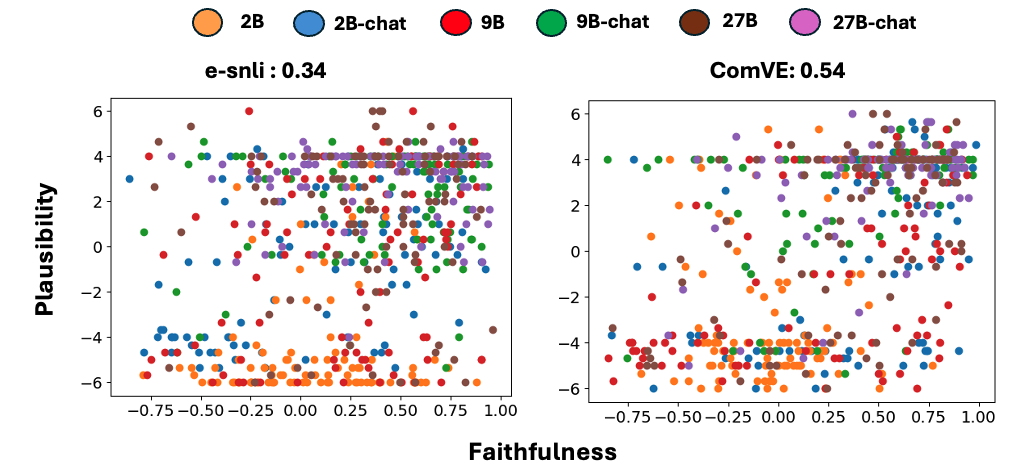}
    \caption{Pearson's correlation between plausibility and faithfulness.}
    \label{fig:plaus}
\end{figure*}

\subsection{Plausibility vs Faithfulness}
\label{sec:plaus}
Ensuring faithful explanations is crucial to prevent over-reliance on plausible but unfaithful NLEs. Thus, it is important to investigate the correlation between plausibility and faithfulness. We evaluate plausibility using the framework from~\citet{chen2023xplainllm}, with OpenAI's GPT4-o\footnote{https://openai.com/index/hello-gpt-4o/} as the evaluator. Each explanation is judged based on six aspects~\citep{hoffman2018metrics} such as clarity, and sufficiency, with more details in ~\ref{sec:appendix_plaus}. We assess both e-SNLI and ComVE, utilizing their gold-standard explanations as references for GPT4-o. Our analysis reveals a positive correlation between plausibility scores and CaF, with ComVE exhibiting a stronger relationship in Figure~\ref{fig:plaus}. This aligns with observations from ~\ref{sec:faithfulness_tests}, where larger models consistently generate higher-quality NLEs, with instruct-tuned models occupying the upper right quadrant of the scatter plot. These results suggest that scaling models and improving general NLP task performance can enhance the explainability of LLM-generated explanations. However, we urge caution in interpreting these findings due to the limited sample size.

\subsection{Distribution of Activation Patching}
In addition to measuring the overall divergence between the distributions of causal values, we also focus on analyzing the internal patterns within these distributions—specifically, the significance of key input elements such as the prediction itself and corrupted tokens.  Our primary focus is on the aggregated effects of corrupted tokens, with positions denoted by $S$, the model’s prediction $\hat{y}_a$, and all the available answer choices. Given that e-SNLI and ComVE consist of only two possible answers, we limit our study to CoS-E, which offers a broader set of answer possibilities. The results, presented in Figure~\ref{fig:analysis}, reveal a consistent trend across all models: they often attribute higher causal importance to corrupted tokens compared to the prediction. This effect is even more pronounced when the model is explaining. Moreover, when aggregating the effects across all answer choices, the resulting overall causal score is significantly higher than that of the prediction alone. This suggests that the model may be reasoning over alternative options when formulating its answer or generating an explanation.

Interestingly, patching multiple layers distributes the causal values from the corrupt tokens to the answer tokens. Overall, we observe no significant differences in causality distribution trends across different model sizes.

\section{Conclusion}
In this work, we introduced a novel faithfulness metric, \textit{Causal Faithfulness}, building on insights from activation patching.  We demonstrated that the faithfulness of a LLM's NLE can be assessed by evaluating the consistency between the causal distributions underlying the explanation and the model's answer. Additionally, we expanded on existing critiques of current tests, particularly highlighting the risks of grounding faithfulness measurements on out-of-distribution inputs. We argue that, to achieve precise faithfulness assessments, the evaluation process must be rigorously examined to avoid drawing misinformed conclusions. We believe our approach is a step toward achieving true faithfulness—a monumental challenge that demands further research into techniques capable of providing deeper insights into the model's internal behavior. Future work could explore integrating mechanistic interpretability approaches~\citep{nanda2023progress} to develop a more robust and reliable framework for assessing faithfulness.

\section{Limitations}
Given the nature of analyzing model internals for assessing faithfulness, one limitation of conducting such a test is the increased computational requirements. CaF requires a total of $T \cdot L$ forward passes, which is higher than existing surface-level tests. However, this requirement is notably smaller than that of CC-SHAP for smaller models, such as the 2B and 9B models, since CC-SHAP scales linearly with $T$, which often exceeds $L$ in such cases. Additionally, CaF avoids the need for approximating attribution values and does not rely on out-of-distribution samples and by design, is a more theoretically sound framework for measuring faithfulness as compared to existing surface-level methods.

Another limitation concerns the process of token replacement, which requires the set of edited tokens to be of equal length to the tokens being replaced, $S$. This constraint is further complicated by the varying tokenization schemes used by different tokenizers. Future research could explore methods to relax this requirement, potentially enabling variable-length replacements.

Lastly, activation patching primarily provides insights starting from the corrupted spans onward, which may limit the scope of the information revealed. We mitigate this limitation by focusing on samples where the rationales are positioned at the beginning. However, the impact of this constraint varies depending on the task structure. For instance, in ~\citet{meng2022locating}, the subject tokens are typically located at the beginning, and in e-SNLI, we target the premise.

\bibliography{custom}

\begin{thebibliography}{34}
\providecommand{\natexlab}[1]{#1}

\bibitem[{Achiam et~al.(2023)Achiam, Adler, Agarwal, Ahmad, Akkaya, Aleman, Almeida, Altenschmidt, Altman, Anadkat et~al.}]{achiam2023gpt}
Josh Achiam, Steven Adler, Sandhini Agarwal, Lama Ahmad, Ilge Akkaya, Florencia~Leoni Aleman, Diogo Almeida, Janko Altenschmidt, Sam Altman, Shyamal Anadkat, et~al. 2023.
\newblock Gpt-4 technical report.
\newblock \emph{arXiv preprint arXiv:2303.08774}.

\bibitem[{Agarwal et~al.(2024)Agarwal, Tanneru, and Lakkaraju}]{agarwal2024faithfulness}
Chirag Agarwal, Sree~Harsha Tanneru, and Himabindu Lakkaraju. 2024.
\newblock Faithfulness vs. plausibility: On the (un) reliability of explanations from large language models.
\newblock \emph{arXiv preprint arXiv:2402.04614}.

\bibitem[{Atanasova et~al.(2023)Atanasova, Camburu, Lioma, Lukasiewicz, Simonsen, and Augenstein}]{atanasova2023faithfulness}
Pepa Atanasova, Oana-Maria Camburu, Christina Lioma, Thomas Lukasiewicz, Jakob~Grue Simonsen, and Isabelle Augenstein. 2023.
\newblock Faithfulness tests for natural language explanations.
\newblock \emph{arXiv preprint arXiv:2305.18029}.

\bibitem[{Cambria et~al.(2023)Cambria, Mao, Chen, Wang, and Ho}]{cambria2023seven}
Erik Cambria, Rui Mao, Melvin Chen, Zhaoxia Wang, and Seng-Beng Ho. 2023.
\newblock Seven pillars for the future of artificial intelligence.
\newblock \emph{IEEE Intelligent Systems}, 38(6):62--69.

\bibitem[{Camburu et~al.(2018)Camburu, Rockt{\"a}schel, Lukasiewicz, and Blunsom}]{camburu2018snli}
Oana-Maria Camburu, Tim Rockt{\"a}schel, Thomas Lukasiewicz, and Phil Blunsom. 2018.
\newblock e-snli: Natural language inference with natural language explanations.
\newblock \emph{Advances in Neural Information Processing Systems}, 31.

\bibitem[{Chen et~al.(2023)Chen, Chen, Gaidhani, Singh, and Sra}]{chen2023xplainllm}
Zichen Chen, Jianda Chen, Mitali Gaidhani, Ambuj Singh, and Misha Sra. 2023.
\newblock Xplainllm: A qa explanation dataset for understanding llm decision-making.
\newblock \emph{arXiv preprint arXiv:2311.08614}.

\bibitem[{DeYoung et~al.(2019)DeYoung, Jain, Rajani, Lehman, Xiong, Socher, and Wallace}]{deyoung2019eraser}
Jay DeYoung, Sarthak Jain, Nazneen~Fatema Rajani, Eric Lehman, Caiming Xiong, Richard Socher, and Byron~C Wallace. 2019.
\newblock Eraser: A benchmark to evaluate rationalized nlp models.
\newblock \emph{arXiv preprint arXiv:1911.03429}.

\bibitem[{Gat et~al.(2023)Gat, Calderon, Feder, Chapanin, Sharma, and Reichart}]{gat2023faithful}
Yair Gat, Nitay Calderon, Amir Feder, Alexander Chapanin, Amit Sharma, and Roi Reichart. 2023.
\newblock Faithful explanations of black-box nlp models using llm-generated counterfactuals.
\newblock \emph{arXiv preprint arXiv:2310.00603}.

\bibitem[{Hase et~al.(2021)Hase, Xie, and Bansal}]{hase2021out}
Peter Hase, Harry Xie, and Mohit Bansal. 2021.
\newblock The out-of-distribution problem in explainability and search methods for feature importance explanations.
\newblock \emph{Advances in neural information processing systems}, 34:3650--3666.

\bibitem[{Hoffman et~al.(2018)Hoffman, Mueller, Klein, and Litman}]{hoffman2018metrics}
Robert~R Hoffman, Shane~T Mueller, Gary Klein, and Jordan Litman. 2018.
\newblock Metrics for explainable ai: Challenges and prospects.
\newblock \emph{arXiv preprint arXiv:1812.04608}.

\bibitem[{Jacovi and Goldberg(2020)}]{jacovi2020towards}
Alon Jacovi and Yoav Goldberg. 2020.
\newblock Towards faithfully interpretable nlp systems: How should we define and evaluate faithfulness?
\newblock \emph{arXiv preprint arXiv:2004.03685}.

\bibitem[{Jacovi and Goldberg(2021)}]{jacovi2021aligning}
Alon Jacovi and Yoav Goldberg. 2021.
\newblock Aligning faithful interpretations with their social attribution.
\newblock \emph{Transactions of the Association for Computational Linguistics}, 9:294--310.

\bibitem[{Jie et~al.(2024)Jie, Satapathy, Goh, and Cambria}]{jie2024interpretable}
Yeo~Wei Jie, Ranjan Satapathy, Rick Goh, and Erik Cambria. 2024.
\newblock How interpretable are reasoning explanations from prompting large language models?
\newblock In \emph{Findings of the Association for Computational Linguistics: NAACL 2024}, pages 2148--2164.

\bibitem[{Lanham et~al.(2023)Lanham, Chen, Radhakrishnan, Steiner, Denison, Hernandez, Li, Durmus, Hubinger, Kernion et~al.}]{lanham2023measuring}
Tamera Lanham, Anna Chen, Ansh Radhakrishnan, Benoit Steiner, Carson Denison, Danny Hernandez, Dustin Li, Esin Durmus, Evan Hubinger, Jackson Kernion, et~al. 2023.
\newblock Measuring faithfulness in chain-of-thought reasoning.
\newblock \emph{arXiv preprint arXiv:2307.13702}.

\bibitem[{Lei et~al.(2016)Lei, Barzilay, and Jaakkola}]{lei2016rationalizing}
Tao Lei, Regina Barzilay, and Tommi Jaakkola. 2016.
\newblock Rationalizing neural predictions.
\newblock \emph{arXiv preprint arXiv:1606.04155}.

\bibitem[{Lundberg and Lee(2017)}]{lundberg2017unified}
Scott~M Lundberg and Su-In Lee. 2017.
\newblock A unified approach to interpreting model predictions.
\newblock \emph{Advances in neural information processing systems}, 30.

\bibitem[{Meng et~al.(2022)Meng, Bau, Andonian, and Belinkov}]{meng2022locating}
Kevin Meng, David Bau, Alex Andonian, and Yonatan Belinkov. 2022.
\newblock Locating and editing factual associations in gpt.
\newblock \emph{Advances in Neural Information Processing Systems}, 35:17359--17372.

\bibitem[{Nanda et~al.(2023)Nanda, Chan, Lieberum, Smith, and Steinhardt}]{nanda2023progress}
Neel Nanda, Lawrence Chan, Tom Lieberum, Jess Smith, and Jacob Steinhardt. 2023.
\newblock Progress measures for grokking via mechanistic interpretability.
\newblock \emph{arXiv preprint arXiv:2301.05217}.

\bibitem[{Parcalabescu and Frank(2023)}]{parcalabescu2023measuring}
Letitia Parcalabescu and Anette Frank. 2023.
\newblock On measuring faithfulness of natural language explanations.
\newblock \emph{arXiv preprint arXiv:2311.07466}.

\bibitem[{Paul et~al.(2024)Paul, West, Bosselut, and Faltings}]{paul2024making}
Debjit Paul, Robert West, Antoine Bosselut, and Boi Faltings. 2024.
\newblock Making reasoning matter: Measuring and improving faithfulness of chain-of-thought reasoning.
\newblock \emph{arXiv preprint arXiv:2402.13950}.

\bibitem[{Pearl(2022)}]{pearl2022direct}
Judea Pearl. 2022.
\newblock Direct and indirect effects.
\newblock In \emph{Probabilistic and causal inference: the works of Judea Pearl}, pages 373--392.

\bibitem[{Rajani et~al.(2019)Rajani, McCann, Xiong, and Socher}]{rajani2019explain}
Nazneen~Fatema Rajani, Bryan McCann, Caiming Xiong, and Richard Socher. 2019.
\newblock Explain yourself! leveraging language models for commonsense reasoning.
\newblock \emph{arXiv preprint arXiv:1906.02361}.

\bibitem[{Siegel et~al.(2024)Siegel, Camburu, Heess, and Perez-Ortiz}]{siegel2024probabilities}
Noah~Y Siegel, Oana-Maria Camburu, Nicolas Heess, and Maria Perez-Ortiz. 2024.
\newblock The probabilities also matter: A more faithful metric for faithfulness of free-text explanations in large language models.
\newblock \emph{arXiv preprint arXiv:2404.03189}.

\bibitem[{Team et~al.(2024)Team, Mesnard, Hardin, Dadashi, Bhupatiraju, Pathak, Sifre, Rivi{\`e}re, Kale, Love et~al.}]{team2024gemma}
Gemma Team, Thomas Mesnard, Cassidy Hardin, Robert Dadashi, Surya Bhupatiraju, Shreya Pathak, Laurent Sifre, Morgane Rivi{\`e}re, Mihir~Sanjay Kale, Juliette Love, et~al. 2024.
\newblock Gemma: Open models based on gemini research and technology.
\newblock \emph{arXiv preprint arXiv:2403.08295}.

\bibitem[{Turpin et~al.(2024)Turpin, Michael, Perez, and Bowman}]{turpin2024language}
Miles Turpin, Julian Michael, Ethan Perez, and Samuel Bowman. 2024.
\newblock Language models don't always say what they think: unfaithful explanations in chain-of-thought prompting.
\newblock \emph{Advances in Neural Information Processing Systems}, 36.

\bibitem[{Vaswani et~al.(2017)Vaswani, Shazeer, Parmar, Uszkoreit, Jones, Gomez, Kaiser, and Polosukhin}]{vaswani2017attention}
Ashish Vaswani, Noam Shazeer, Niki Parmar, Jakob Uszkoreit, Llion Jones, Aidan~N Gomez, {\L}ukasz Kaiser, and Illia Polosukhin. 2017.
\newblock Attention is all you need.
\newblock \emph{Advances in neural information processing systems}, 30.

\bibitem[{Vig et~al.(2020)Vig, Gehrmann, Belinkov, Qian, Nevo, Sakenis, Huang, Singer, and Shieber}]{vig2020causal}
Jesse Vig, Sebastian Gehrmann, Yonatan Belinkov, Sharon Qian, Daniel Nevo, Simas Sakenis, Jason Huang, Yaron Singer, and Stuart Shieber. 2020.
\newblock Causal mediation analysis for interpreting neural nlp: The case of gender bias.
\newblock \emph{arXiv preprint arXiv:2004.12265}.

\bibitem[{Wang et~al.(2020)Wang, Liang, Jin, Wang, Zhu, and Zhang}]{wang2020semeval}
Cunxiang Wang, Shuailong Liang, Yili Jin, Yilong Wang, Xiaodan Zhu, and Yue Zhang. 2020.
\newblock Semeval-2020 task 4: Commonsense validation and explanation.
\newblock \emph{arXiv preprint arXiv:2007.00236}.

\bibitem[{Wei et~al.(2022)Wei, Wang, Schuurmans, Bosma, Xia, Chi, Le, Zhou et~al.}]{wei2022chain}
Jason Wei, Xuezhi Wang, Dale Schuurmans, Maarten Bosma, Fei Xia, Ed~Chi, Quoc~V Le, Denny Zhou, et~al. 2022.
\newblock Chain-of-thought prompting elicits reasoning in large language models.
\newblock \emph{Advances in neural information processing systems}, 35:24824--24837.

\bibitem[{Wei~Jie et~al.(2024)Wei~Jie, Satapathy, and Cambria}]{jie2024plausible}
Yeo Wei~Jie, Ranjan Satapathy, and Erik Cambria. 2024.
\newblock \href {https://aclanthology.org/2024.findings-acl.307} {Plausible extractive rationalization through semi-supervised entailment signal}.
\newblock In \emph{Findings of the Association for Computational Linguistics ACL 2024}, pages 5182--5192. Association for Computational Linguistics.

\bibitem[{Wiegreffe et~al.(2020)Wiegreffe, Marasovi{\'c}, and Smith}]{wiegreffe2020measuring}
Sarah Wiegreffe, Ana Marasovi{\'c}, and Noah~A Smith. 2020.
\newblock Measuring association between labels and free-text rationales.
\newblock \emph{arXiv preprint arXiv:2010.12762}.

\bibitem[{Yeo et~al.(2024)Yeo, Ferdinan, Kazienko, Satapathy, and Cambria}]{jie2024selftraining}
Wei~Jie Yeo, Teddy Ferdinan, Przemyslaw Kazienko, Ranjan Satapathy, and Erik Cambria. 2024.
\newblock \href {https://arxiv.org/abs/2406.11275} {Self-training large language models through knowledge detection}.
\newblock \emph{Preprint}, arXiv:2406.11275.

\bibitem[{Yeo et~al.(2023)Yeo, van~der Heever, Mao, Cambria, Satapathy, and Mengaldo}]{yeo2023comprehensive}
Wei~Jie Yeo, Wihan van~der Heever, Rui Mao, Erik Cambria, Ranjan Satapathy, and Gianmarco Mengaldo. 2023.
\newblock A comprehensive review on financial explainable ai.
\newblock \emph{arXiv preprint arXiv:2309.11960}.

\bibitem[{Zhang and Nanda(2023)}]{zhang2023towards}
Fred Zhang and Neel Nanda. 2023.
\newblock Towards best practices of activation patching in language models: Metrics and methods.
\newblock \emph{arXiv preprint arXiv:2309.16042}.

\end{thebibliography}

\appendix

\section{Appendix}
\label{sec:appendix}
\subsection{CaF implementation}
\label{sec:implementations}
In this section, we detail the implementation of token replacement for CaF. In CoS-E and e-SNLI, we are provided with annotated indices indicating the important positions within the input, $S = {1, 2, \ldots, M}$, where $M$ varies across samples. We extract the corresponding tokens, referred to as the rationale, and manually edit them to transform the original answer, $y$, into a counterfactual answer, $y^C \neq y$. For CoS-E, a 5-way multiple-choice QA dataset, we select the most plausible alternative answer and modify the rationale accordingly. In e-SNLI, we exclude samples labeled as neutral, as they are not annotated with $S$, and modify the rationale to switch entailment inputs to contradiction and vice versa. We target task A of ComVE, where the model is given two sentences and must identify the statement that contradicts common sense, we adapt the task by presenting the model with only the illogical statement and asking it to confirm its illogicality. We then replace the statement with the logical one to serve as the counterfactual.

We sample both the answer and explanation using a fixed temperature of $1.0$ and nucleus sampling with a threshold of $0.95$. One advantage of performing STR over GN is the reduced batch size, as~\citet{meng2022locating} averages causal effects across 10 different noise perturbations at each layer and feature level, whereas STR requires only a single forward pass per patching. Furthermore, unlike CC-SHAP, which relies on Monte Carlo sampling to handle its exponential computational demands, CaF avoids the need for such approximation techniques.

\begin{figure*}[ht]
    \centering
    \begin{subfigure}[b]{0.45\textwidth}
        \centering
        \includegraphics[width=\textwidth]{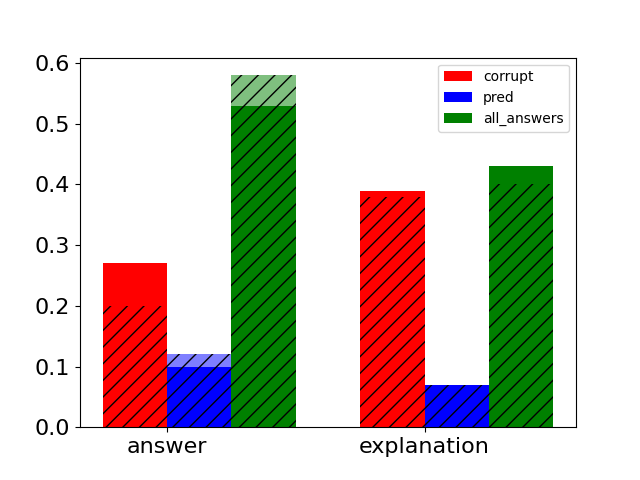}
        \caption{Gemma2-2B}
        \label{fig:2b}
    \end{subfigure}
    \begin{subfigure}[b]{0.45\textwidth}
        \centering
        \includegraphics[width=\textwidth]{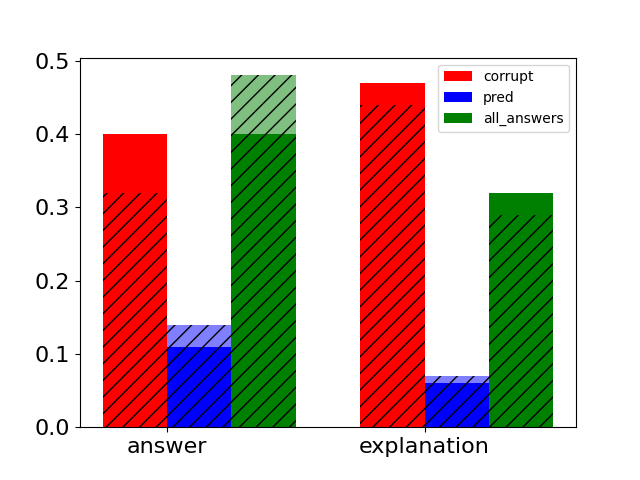}
        \caption{Gemma2-2B-chat}
        \label{fig:2b-chat}
    \end{subfigure}
    \begin{subfigure}[b]{0.45\textwidth}
        \centering
        \includegraphics[width=\textwidth]{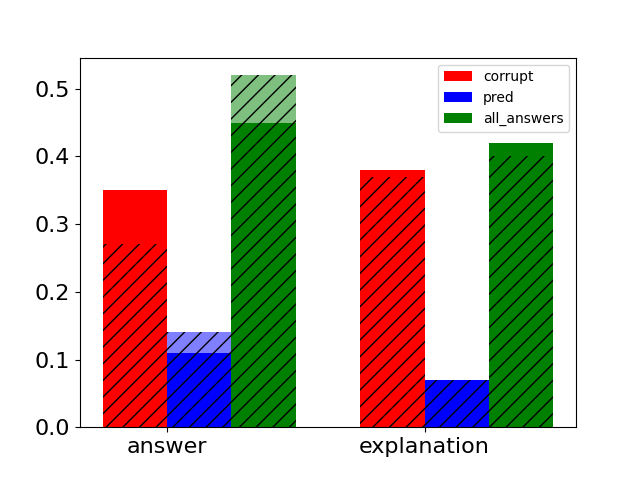}
        \caption{Gemma2-9B}
        \label{fig:9b}
    \end{subfigure}
    \begin{subfigure}[b]{0.45\textwidth}
        \centering
        \includegraphics[width=\textwidth]{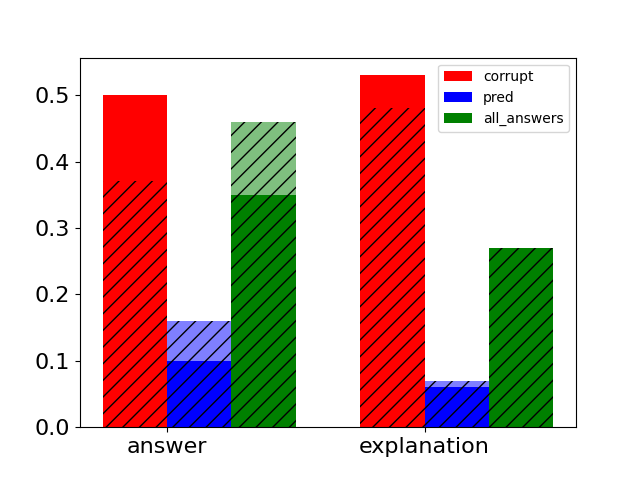}
        \caption{Gemma2-9B-chat}
        \label{fig:9b-chat}
    \end{subfigure}
    \begin{subfigure}[b]{0.45\textwidth}
        \centering
        \includegraphics[width=\textwidth]{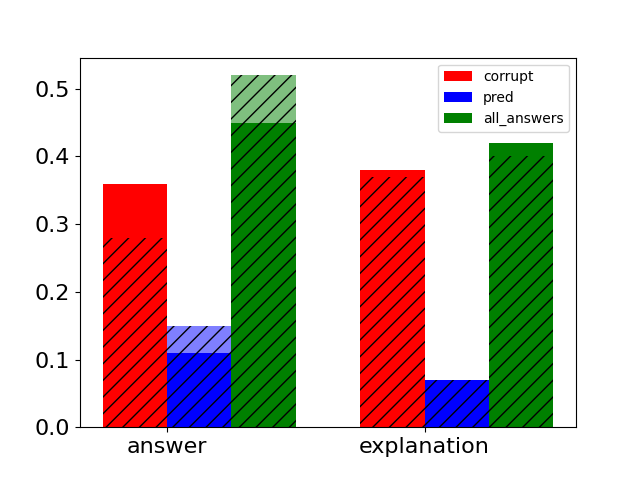}
        \caption{Gemma2-27B}
        \label{fig:27b}
    \end{subfigure}
    \begin{subfigure}[b]{0.45\textwidth}
        \centering
        \includegraphics[width=\textwidth]{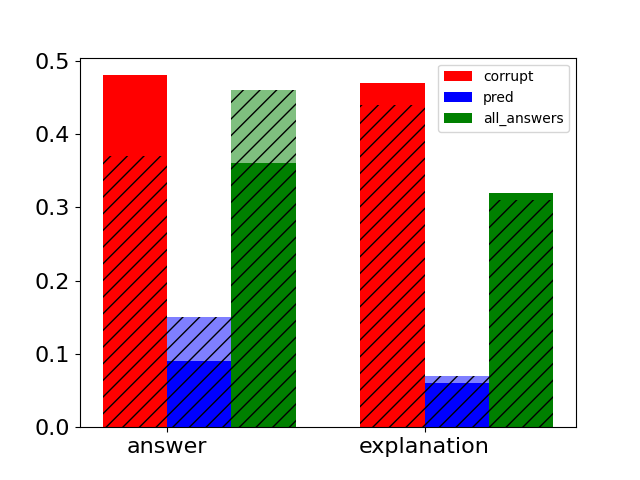}
        \caption{Gemma2-27B-chat}
        \label{fig:27b-chat}
    \end{subfigure}
    \caption{Causal scores at the token level, CaF(T) across the six models on CoS-E. The cross-lines refer to patching multiple layers with window of size $10$. Each bar represents the aggregated values of the target features, red: the corrupted token spanned by $S$, blue: the answer choice corresponding to the resultant prediction and green: all answer choices.}
    \label{fig:analysis}
\end{figure*}

\subsection{Plausibility}
\label{sec:appendix_plaus}
In this section, we discuss the evaluation framework to assess the plausibility criteria of each NLE. We follow the framework used introduced in ~\citet{chen2023xplainllm}, which poses seven questions to the evaluator. We left out the question regarding trust as we find it unrelated to plausibility. Each question is allocated a score using a three-point Likert scale: $-1$ (disagree), $0$ (neutral), and $1$ (agree). For each sample, we prompted the evaluator thrice and averaged the scores to represent a single evaluation. Each plausibility score is thus bounded between $[-6,6]$. While concerns may arise over using LLMs to judge plausibility, the strong correlation between human and model preferences reported in~\citet{chen2023xplainllm} provides sufficient justification for relying on LLM-based evaluations. The prompt template can be shown in Table~\ref{tab:plaus_prompt} 
\clearpage

\begin{table*}[ht]
\centering
\begin{tabular}{p{0.9\textwidth}}
\toprule
Please rate the plausibility and quality of the candidate explanation generated by a langauge model to support its own answer to the corresponding question. Assign a score of either -1 for disagree, 0 for neutral, or 1 for agree by answering the following criteria questions below. You are also given a list of gold explanation samples for reference. \\
Question: <question> \\
Choices: <choices> \\
Answer: <answer> \\
Candidate Explanation: <explanation> \\
Gold Explanation Samples: <gold explanations> \\\\
Criteria Questions: \\
Q1: This is a good explanation \\
1. Disagree: The explanation is illogical or inconsistent with the question and/or does not adequately cover the answer choices \\
2. Neutral: The explanation is somewhat logical and consistent with the question but might miss some aspects of the answer choices. \\
3. Agree: The explanation is logical, consistent with the question, and adequately covers the answer choices. \\

Q2: I understand this explanation of how the AI model works. \\
1. Disagree: The explanation is unclear or contains overly complex terms or convoluted sentences. \\
2. Neutral: The explanation is somewhat understandable but might contain complex terms or convoluted sentences. \\
3. Agree: The explanation is clear, concise, and easy to understand. \\

Q3: This explanation of how the AI model works is satisfying.\\
1. Disagree: The explanation does not meet my expectations and leaves many questions unanswered.\\
2. Neutral: The explanation somewhat meets my expectations but leaves some questions unanswered.\\
3. Agree: The explanation meets my expectations and satisfies my query.\\

Q4: This explanation of how the AI model works has sufficient detail.\\
1. Disagree: The explanation lacks detail and does not adequately cover the AI model’s decisionmaking.\\
2. Neutral: The explanation provides some detail but lacks thoroughness in covering the AI model’s decision-making.\\
3. Agree: The explanation is thorough and covers all aspects of the AI model’s decision-making.\\

Q5: This explanation of how the AI model works seems complete.\\
1. Disagree: The explanation does not adequately cover the answer choices and leaves many aspects unexplained.\\
2. Neutral: The explanation covers most answer choices but leaves some aspects unexplained.\\
3. Agree: The explanation covers all answer choices and leaves no aspect unexplained.\\

Q6: This explanation of how the AI model works is accurate.\\
1. Disagree: The explanation does not accurately reflect the AI model’s decision-making.\\
2. Neutral: The explanation somewhat reflects the AI model’s decision-making but contains some inaccuracies.\\
3. Agree: The explanation accurately reflects the AI model’s decision-making. \\
\bottomrule
\end{tabular}
\caption{Prompt template for plausibility scoring.}
\label{tab:plaus_prompt}
\end{table*}
\clearpage

\section{Examples of CaF}
In this section, we illustrate the causal matrix, $C_a$ and $C_e$ on an example for each dataset: CoS-E, e-snli and ComVE for all six models. Each matrix is truncated from the corrupted span onwards, and we include the counterfactual edits used by token replacement. We omit the values since CaF only evaluates the consistency between the distributions. 
\newpage

\begin{figure*}[ht]
    \centering
    \begin{subfigure}[b]{\textwidth}
        \centering
        \includegraphics[width=\textwidth]{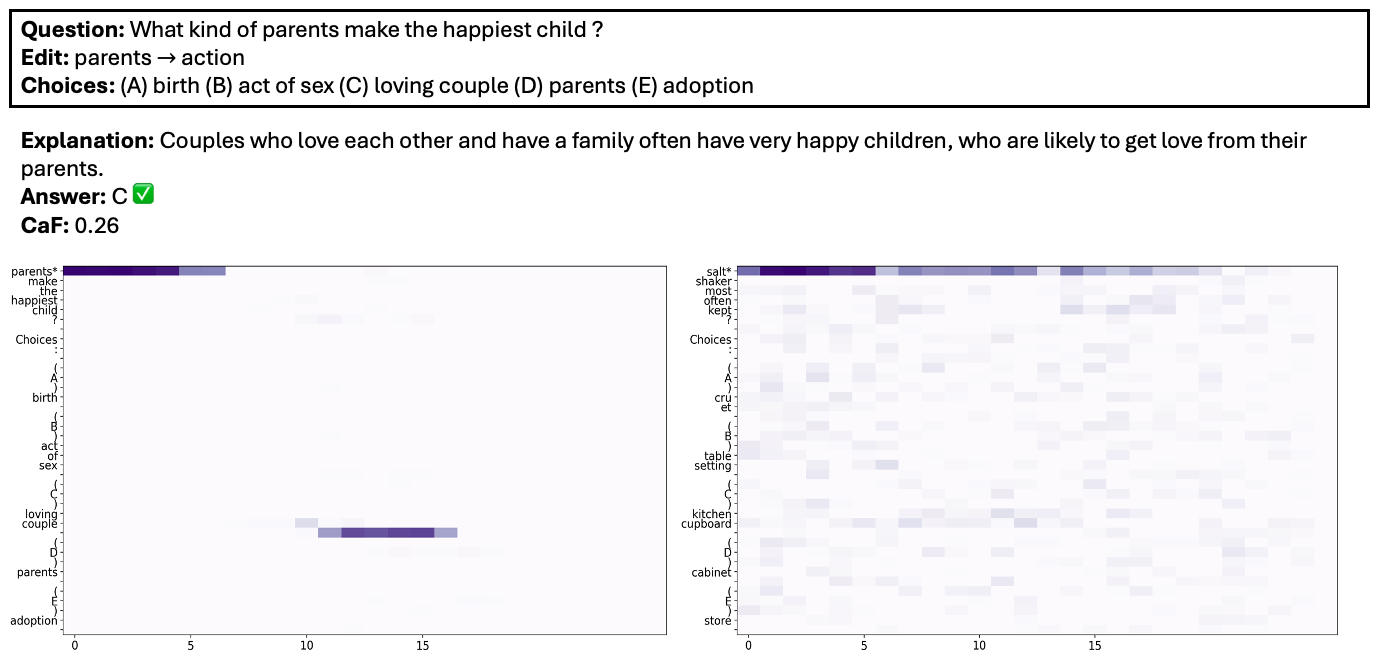}
        \caption{Gemma2-2B}
    \end{subfigure}
    \begin{subfigure}[b]{\textwidth}
        \centering
        \includegraphics[width=\textwidth]{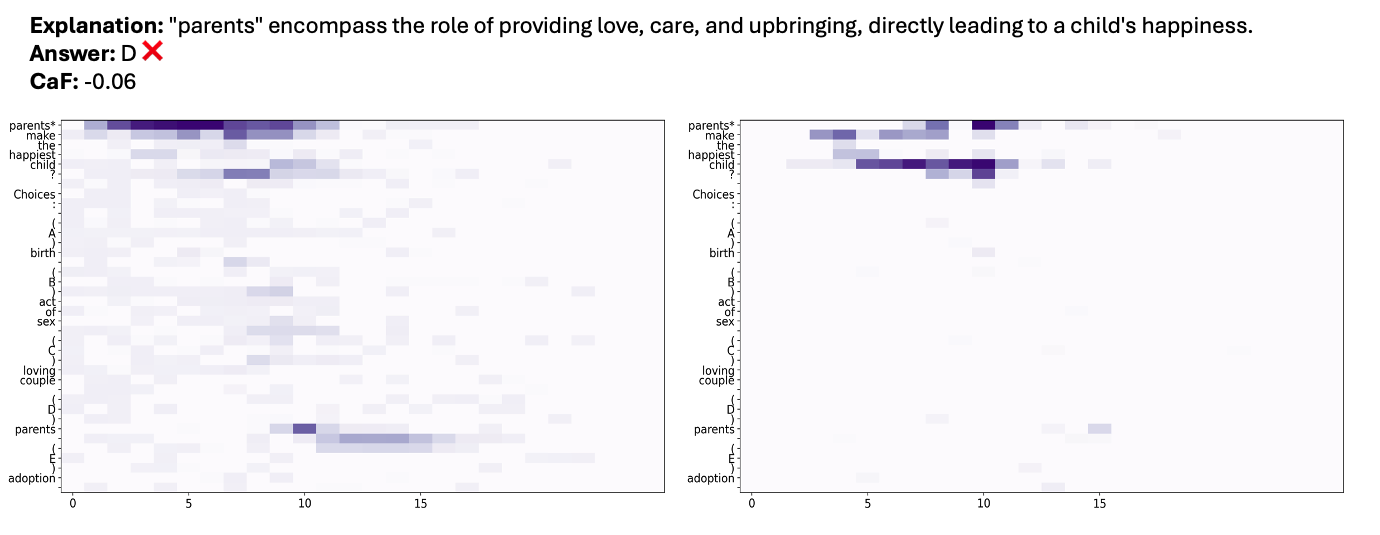}
        \caption{Gemma2-2B-chat}
    \end{subfigure}
    \caption{Illustration of causal attributions on CoS-E, between Gemma2-2B pretrained [top] and instruct-tuned [bottom] LLM. Left refers to the values pertaining to the answer generation, $C_a$, and right to the explanation, $C_e$. Both examples show high values on the corresponding prediction choice. The chat model predicted an incorrect answer, focusing on the option D, while the pre-trained model correctly identified "loving couple" as the answer.}
    \label{fig:cose_2b}
\end{figure*}
\newpage


\begin{figure*}[ht]
    \centering
    \begin{subfigure}[b]{\textwidth}
        \centering
        \includegraphics[width=\textwidth]{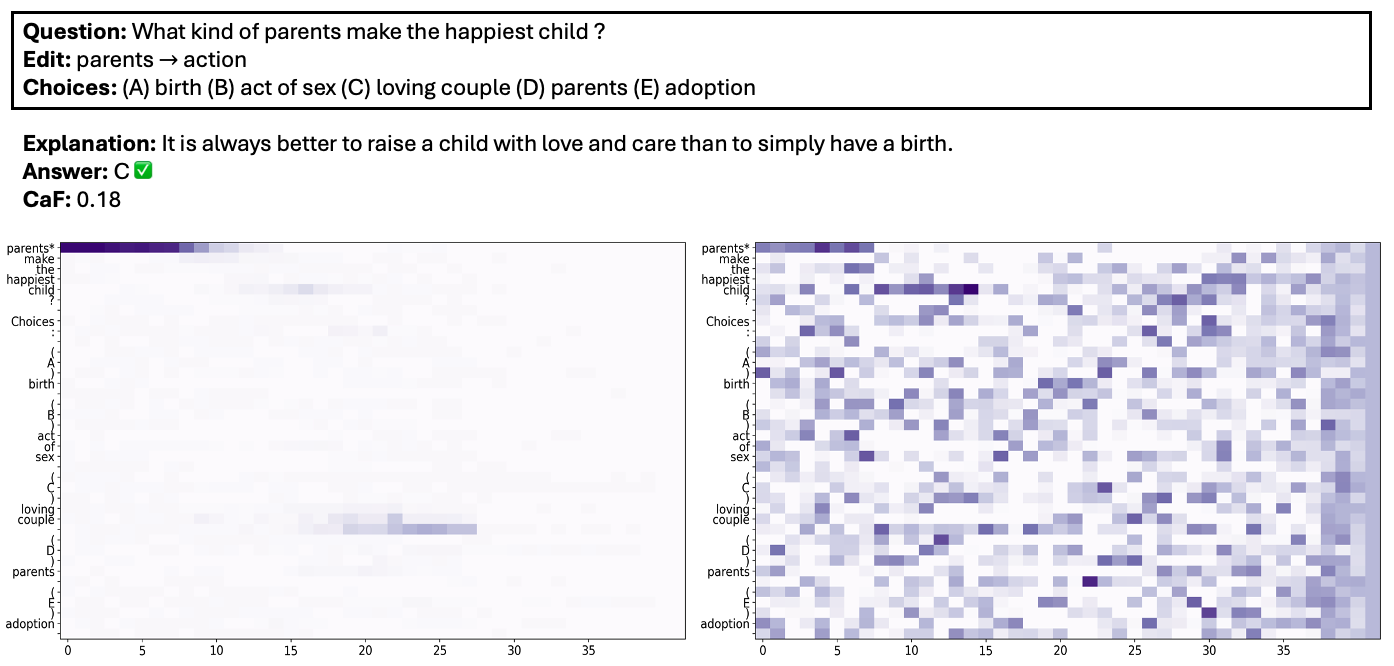}
        \caption{Gemma2-9B}
    \end{subfigure}
    \begin{subfigure}[b]{\textwidth}
        \centering
        \includegraphics[width=\textwidth]{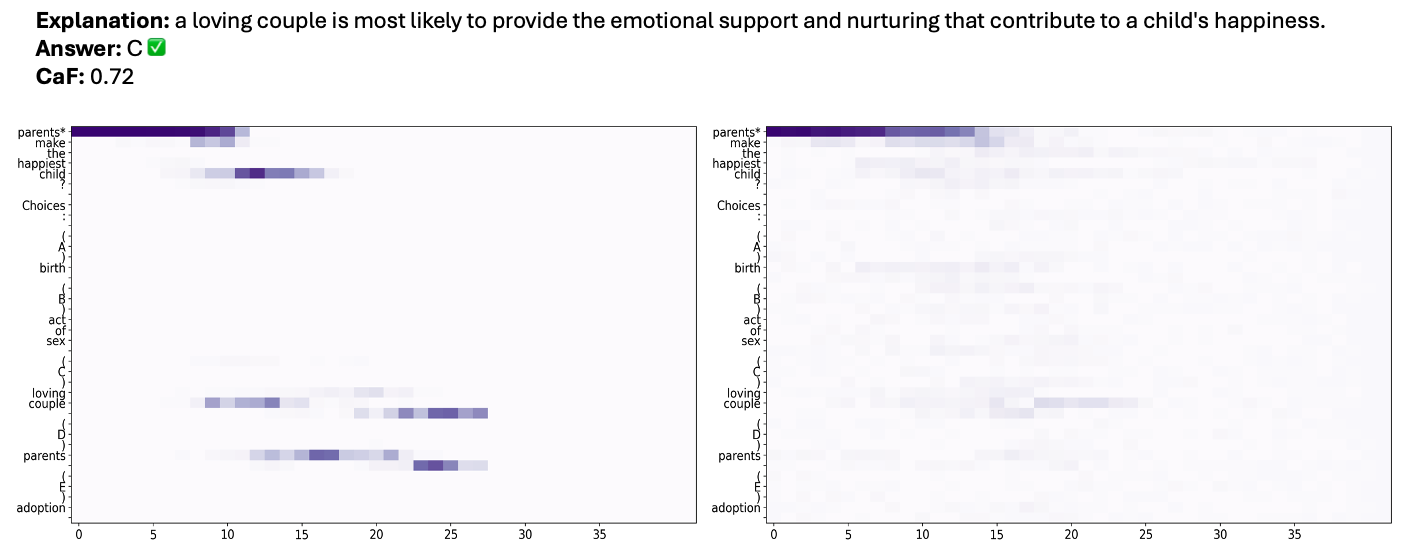}
        \caption{Gemma2-9B-chat}
    \end{subfigure}
    \caption{CoS-E on Gemma2-9B models. The explanation causal values appear to be noisy for the pre-trained model as compared to the chat model, yielding a lower faithfulness score. The chat model reasons between two possible answers, and picks the correct answer, in contrast with ~\ref{fig:cose_2b}, where the chat model only focuses on the wrong answer.}
    \label{fig:cose_9b}
\end{figure*}
\newpage

\begin{figure*}[ht]
    \centering
    \begin{subfigure}[b]{\textwidth}
        \centering
        \includegraphics[width=\textwidth]{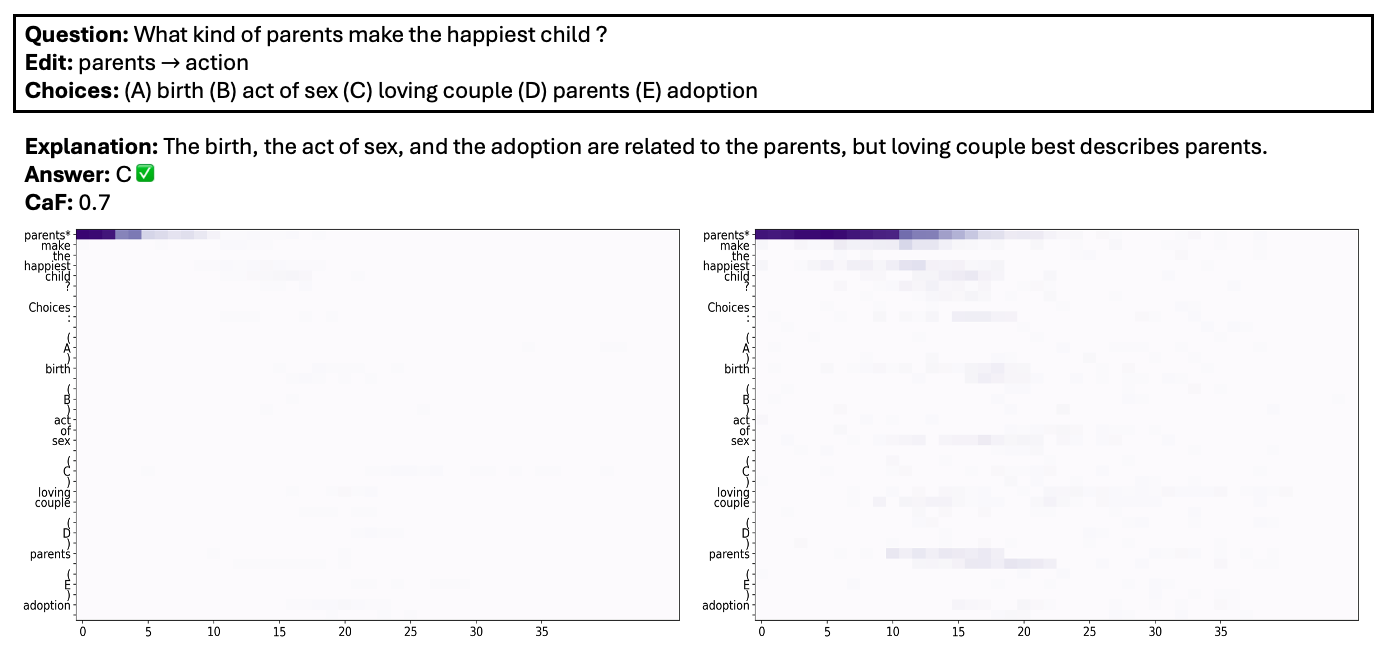}
        \caption{Gemma2-27B}
    \end{subfigure}
    \begin{subfigure}[b]{\textwidth}
        \centering
        \includegraphics[width=\textwidth]{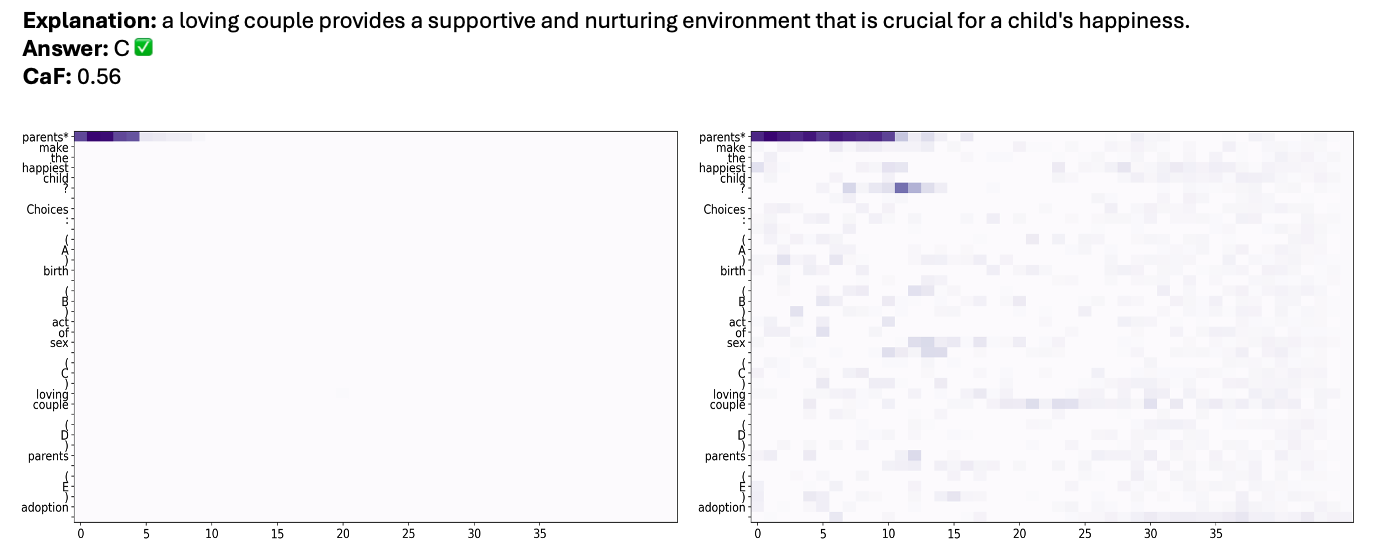}
        \caption{Gemma2-27B-chat}
    \end{subfigure}
    \caption{CoS-E on Gemma2-27B models. Both models attributes higher causality in the earlier layers of the corrupted span.}
    \label{fig:cose_27b}
\end{figure*}
\newpage


\begin{figure*}[ht]
    \centering
    \begin{subfigure}[b]{\textwidth}
        \centering
        \includegraphics[width=\textwidth]{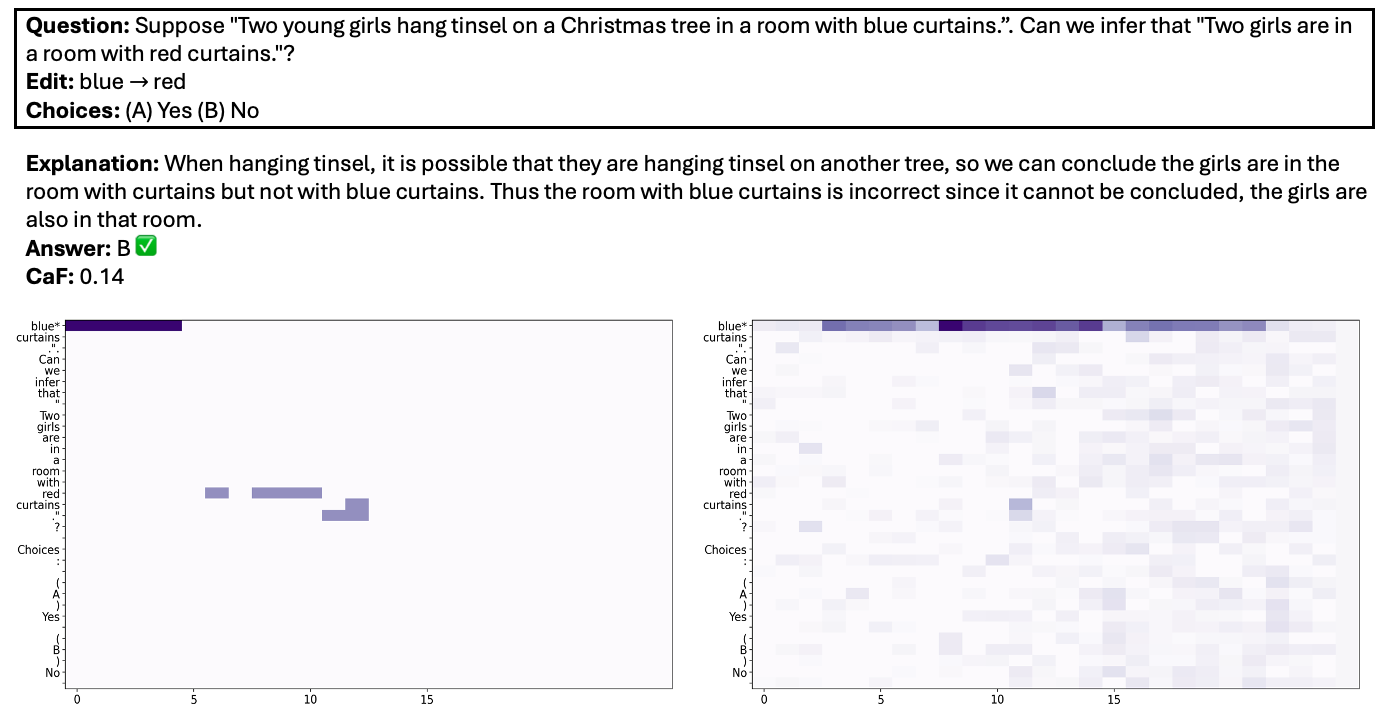}
        \caption{Gemma2-2B}
    \end{subfigure}
    \begin{subfigure}[b]{\textwidth}
        \centering
        \includegraphics[width=\textwidth]{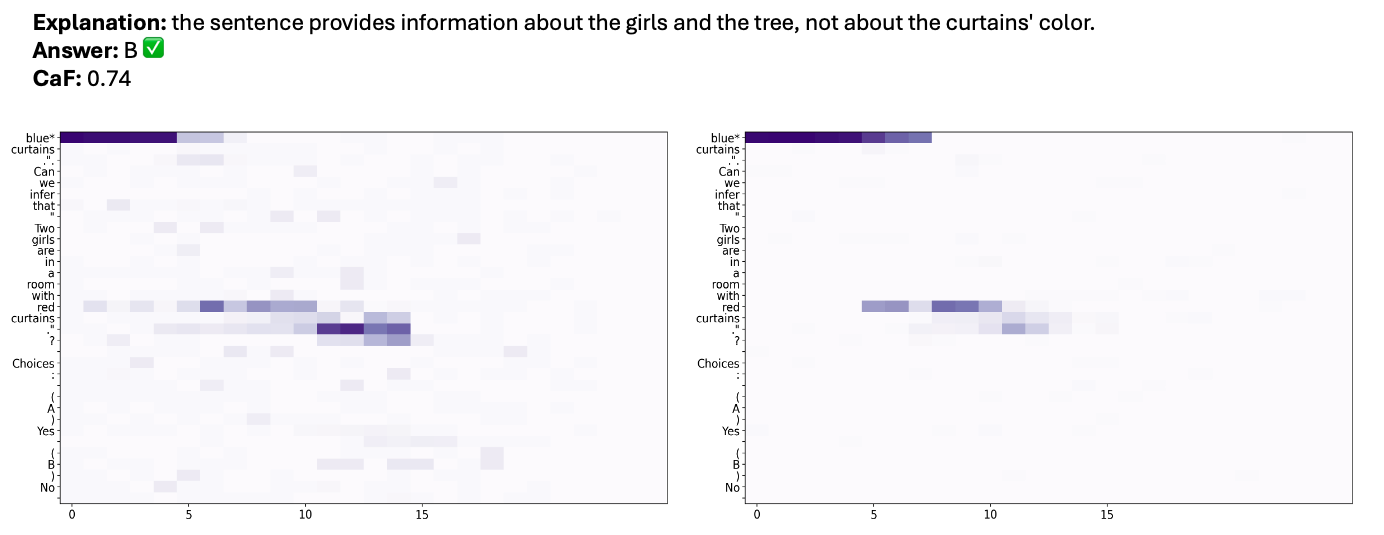}
        \caption{Gemma2-2B-chat}
    \end{subfigure}
    \caption{E-snli: Gemma2-2B LLMs. Both models focus on the corrupted token \textit{"blue"}, but differs in the proceeding tokens. Both models $C_a$ highlight the noun phrase (NP), \textit{"red curtains"} in the hypothesis, attributing it to the contradiction label. However, the $C_e$ in the pre-trained model is noisier as compared to the chat model which appears to be concentrated in the NP.}
    \label{fig:esnli_2b}
\end{figure*}
\newpage

\begin{figure*}[ht]
    \centering
    \begin{subfigure}[b]{\textwidth}
        \centering
        \includegraphics[width=\textwidth]{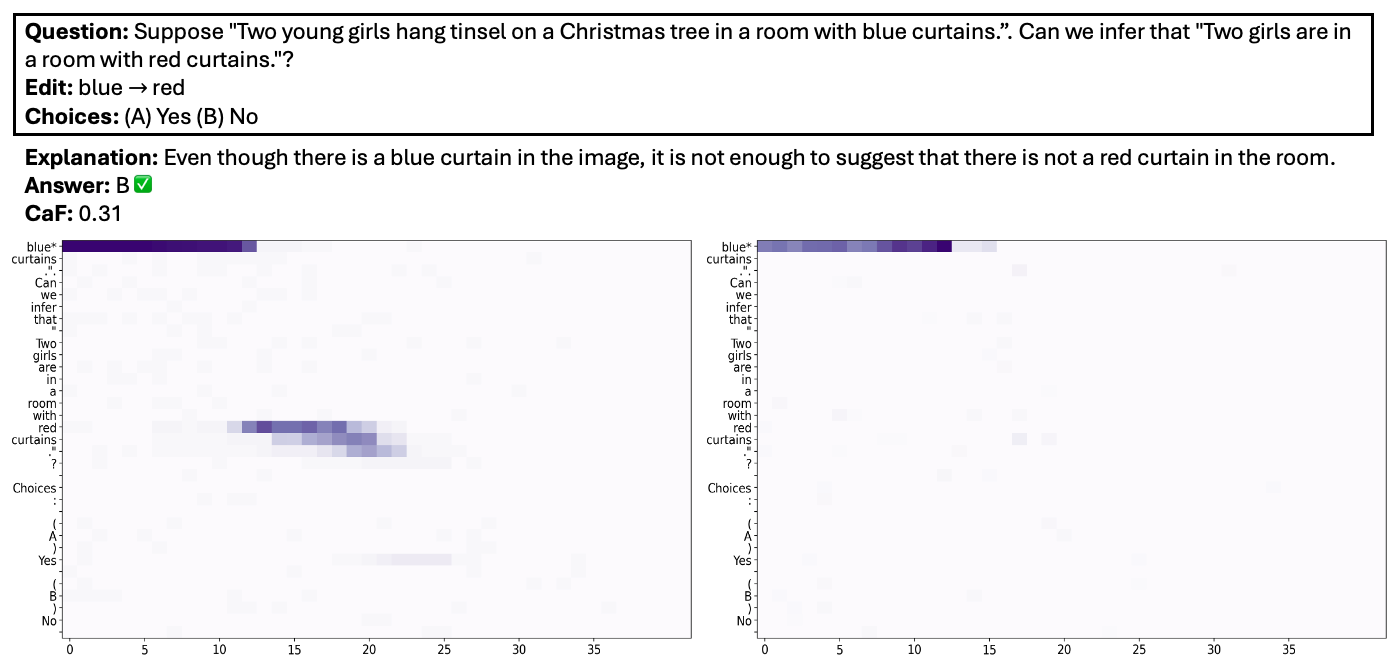}
        \caption{Gemma2-9B}
    \end{subfigure}
    \begin{subfigure}[b]{\textwidth}
        \centering
        \includegraphics[width=\textwidth]{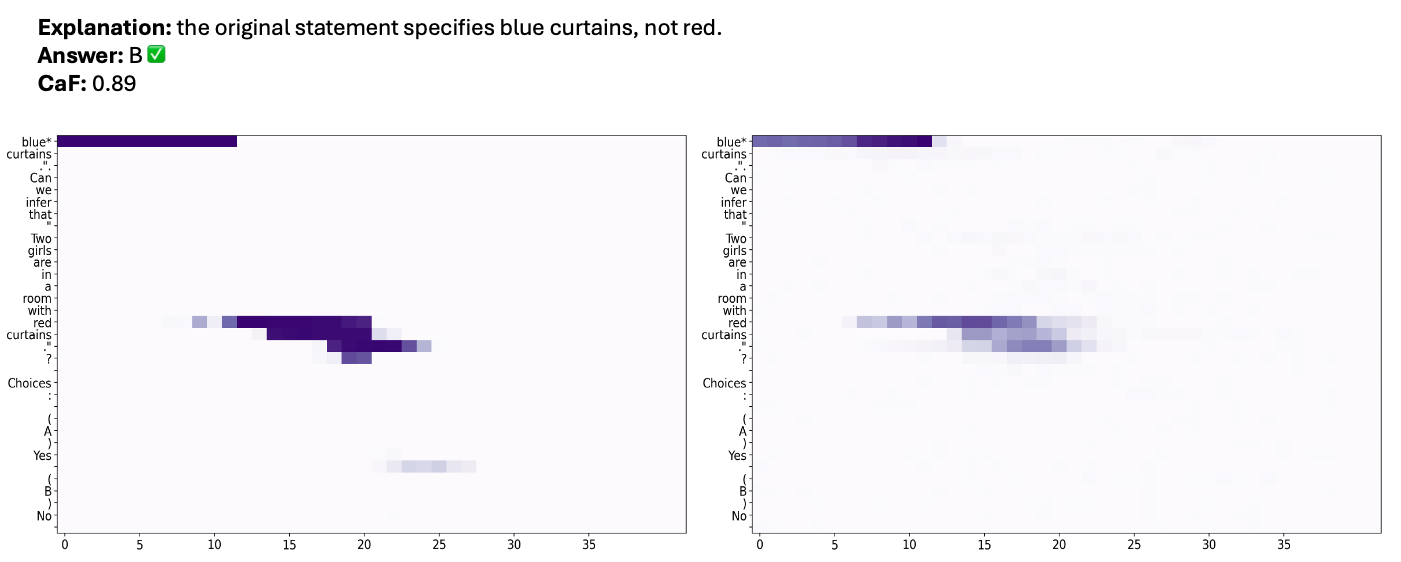}
        \caption{Gemma2-9B-chat}
    \end{subfigure}
    \caption{E-snli: Gemma2-9B LLMs. The 9B models display similar patterns to those in Figure~\ref{fig:esnli_2b}, by focusing primarily on the corrupted token and NP. However, the pre-trained model is less faithful, as the $C_e$ matrix fails to highlight the NP, in contrast to $C_a$. Additionally, the models appear to rely more heavily on the middle layers when reasoning about the NP.}
    \label{fig:esnli_9b}
\end{figure*}
\newpage

\begin{figure*}[ht]
    \centering
    \begin{subfigure}[b]{\textwidth}
        \centering
        \includegraphics[width=\textwidth]{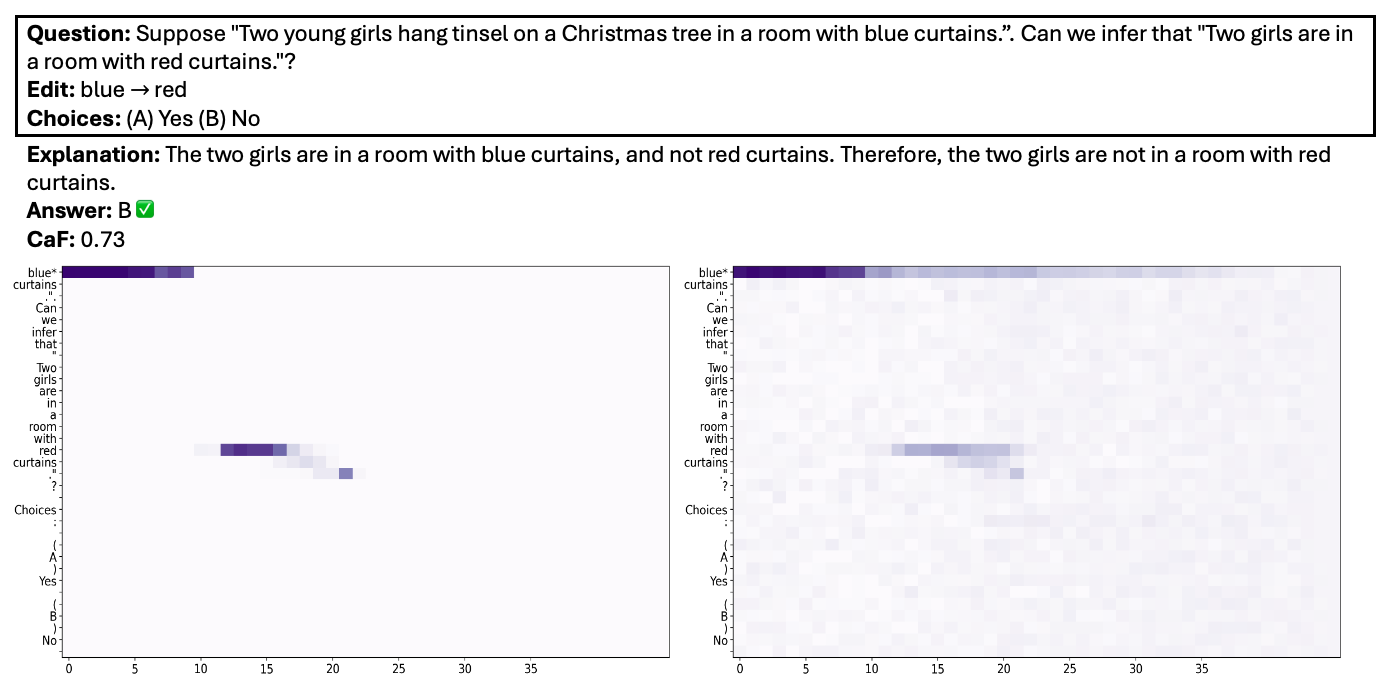}
        \caption{Gemma2-27B}
    \end{subfigure}
    \begin{subfigure}[b]{\textwidth}
        \centering
        \includegraphics[width=\textwidth]{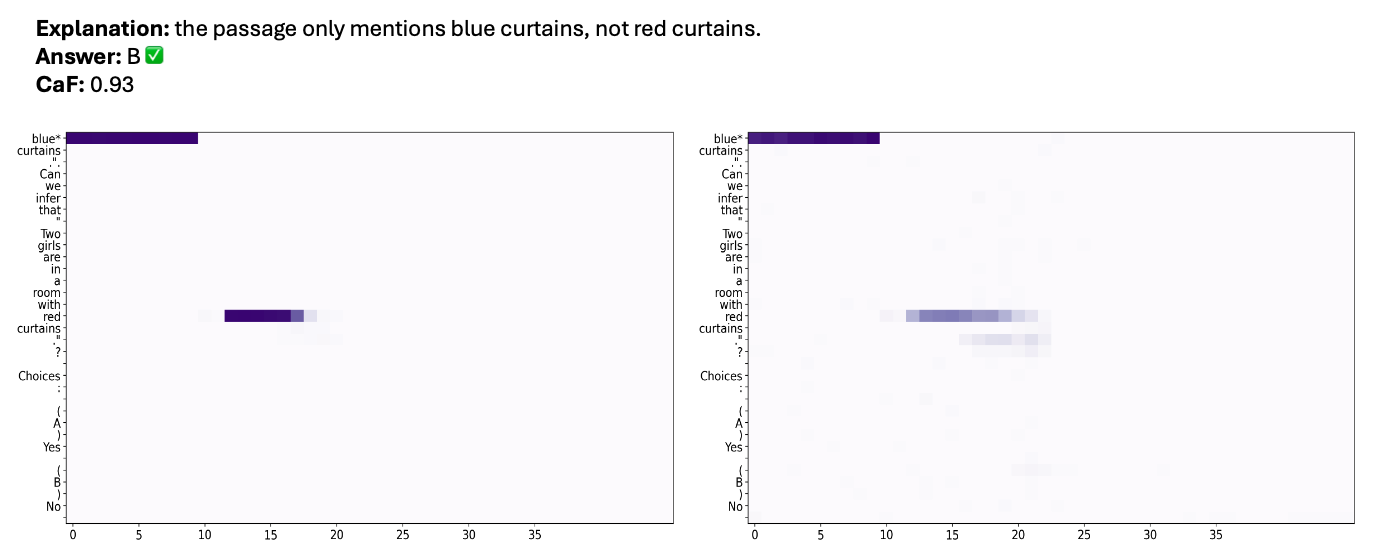}
        \caption{Gemma2-27B-chat}
    \end{subfigure}
    \caption{E-snli: Gemma2-9B LLMs. In contrast to the 2B and 9B models. The causal distributions of $C_a$ and $C_e$ are more consistent, specifically for the pre-trained variant.}
    \label{fig:esnli_27b}
\end{figure*}
\newpage


\begin{figure*}[ht]
    \centering
    \begin{subfigure}[b]{\textwidth}
        \centering
        \includegraphics[width=\textwidth]{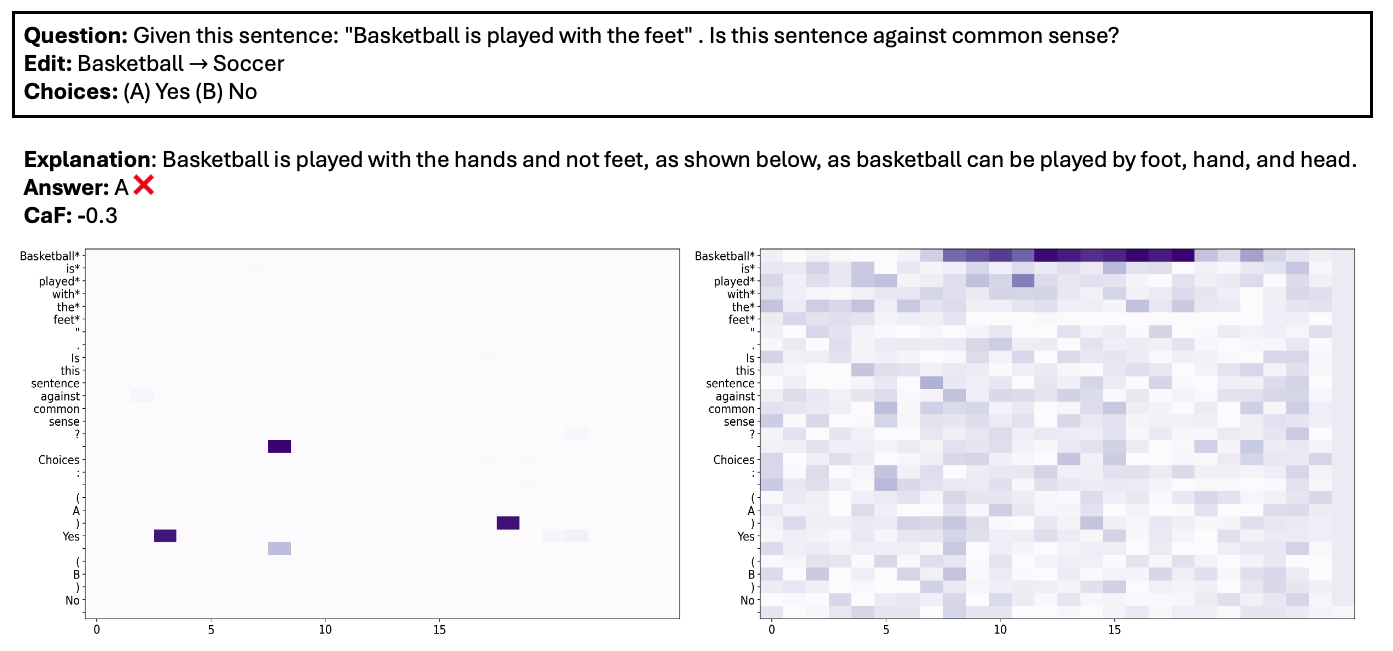}
        \caption{Gemma2-2B}
    \end{subfigure}
    \begin{subfigure}[b]{\textwidth}
        \centering
        \includegraphics[width=\textwidth]{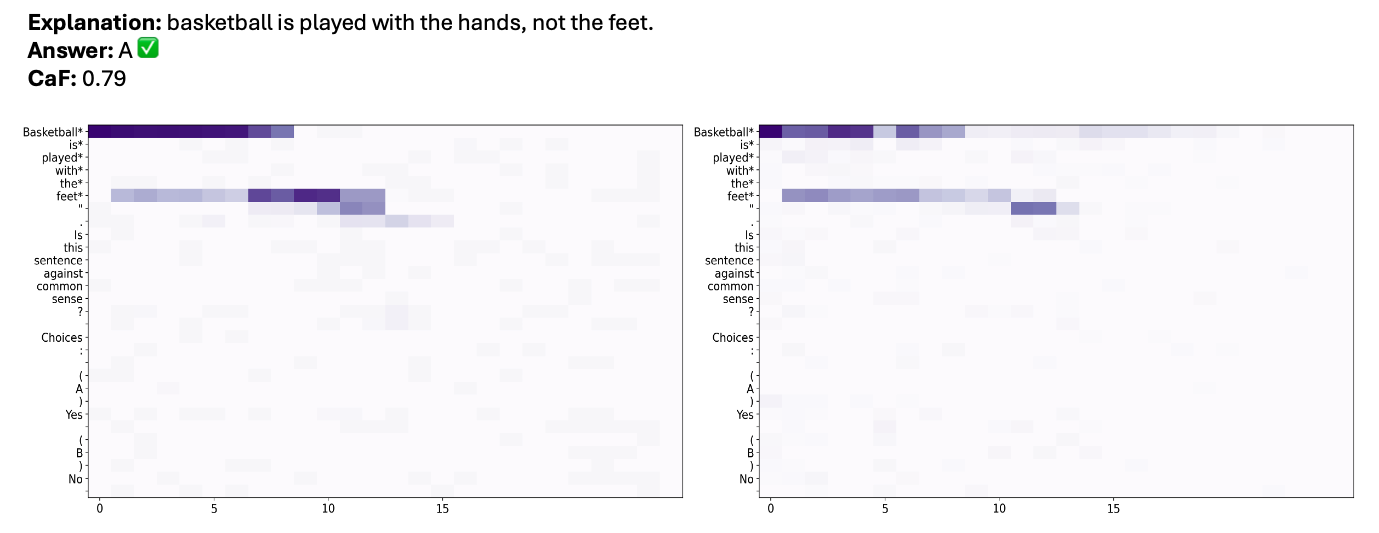}
        \caption{Gemma2-2B-chat}
    \end{subfigure}
    \caption{ComVE: Gemma2-2B LLMs. The pre-trained model is considered unfaithful due to the noisy causal distribution in the explanation, $C_e$, a pattern observed across other datasets. In contrast, the chat model demonstrates a much more consistent distribution between the explanation and the answer. The pre-trained model fails to focus on the key noun, \textit{"Basketball"}, which accounts for the incorrect prediction. This highlights how AP can be used to explain why the model made an erroneous decision, as the pre-trained model concentrates on the wrong elements while neglecting the key token, unlike the chat model.}
    \label{fig:esnli_2b}
\end{figure*}
\newpage

\begin{figure*}[ht]
    \centering
    \begin{subfigure}[b]{\textwidth}
        \centering
        \includegraphics[width=\textwidth]{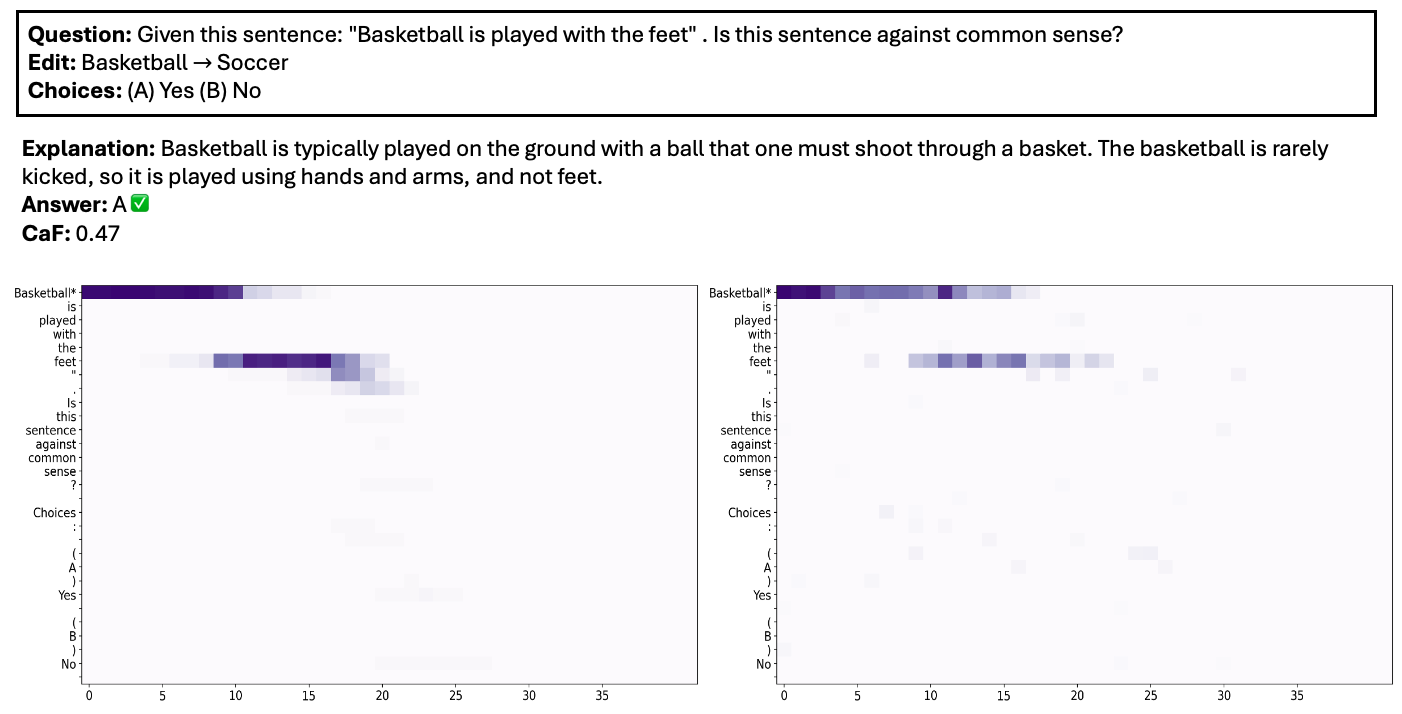}
        \caption{Gemma2-9B}
    \end{subfigure}
    \begin{subfigure}[b]{\textwidth}
        \centering
        \includegraphics[width=\textwidth]{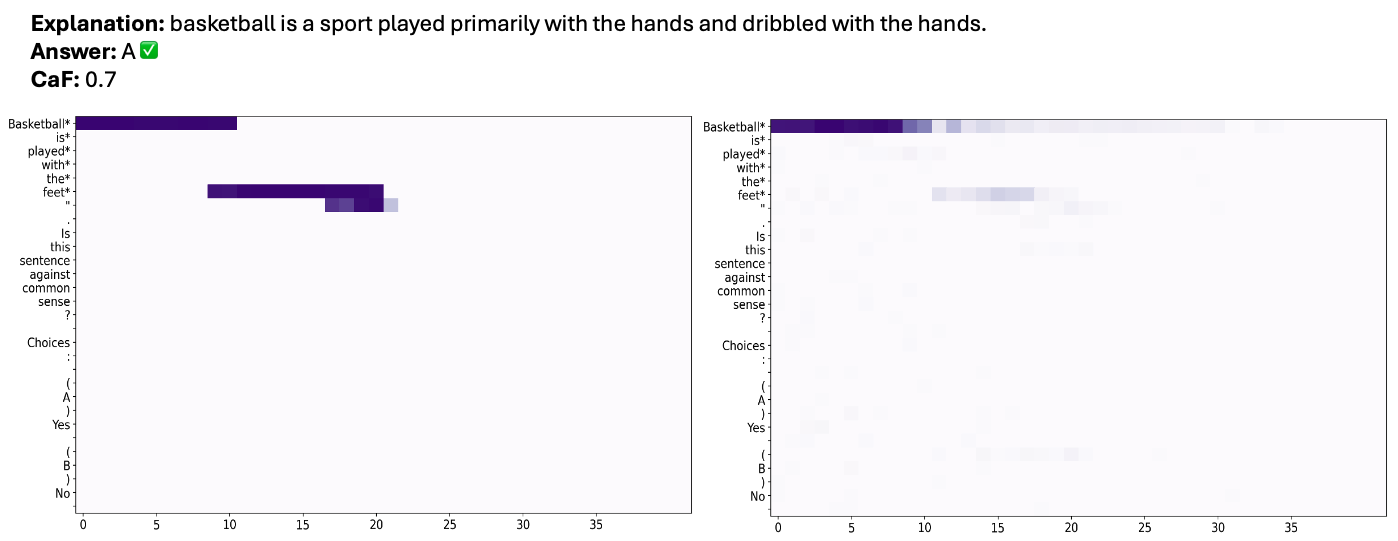}
        \caption{Gemma2-9B-chat}
    \end{subfigure}
    \caption{ComVE: Gemma2-9B LLMs. Both models are relatively faithful, with high values assigned to both the subject and object noun.}
    \label{fig:esnli_9b}
\end{figure*}
\newpage

\begin{figure*}[ht]
    \centering
    \begin{subfigure}[b]{\textwidth}
        \centering
        \includegraphics[width=\textwidth]{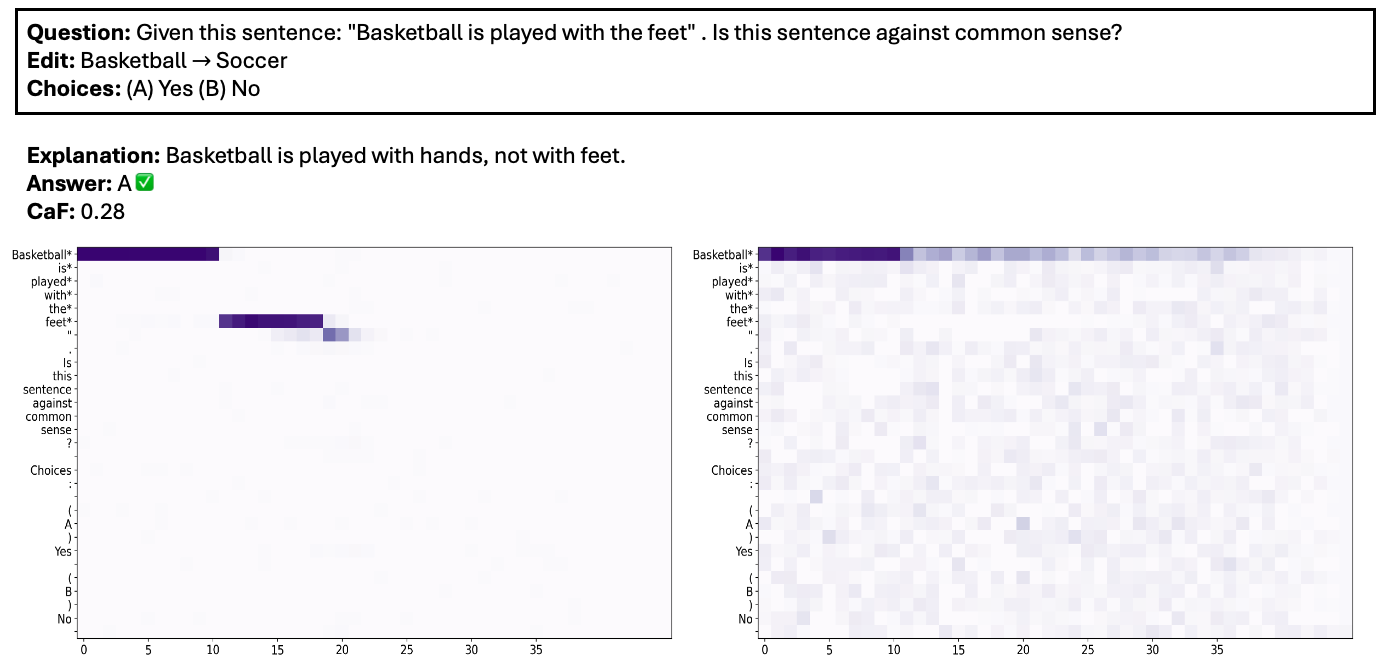}
        \caption{Gemma2-27B}
    \end{subfigure}
    \begin{subfigure}[b]{\textwidth}
        \centering
        \includegraphics[width=\textwidth]{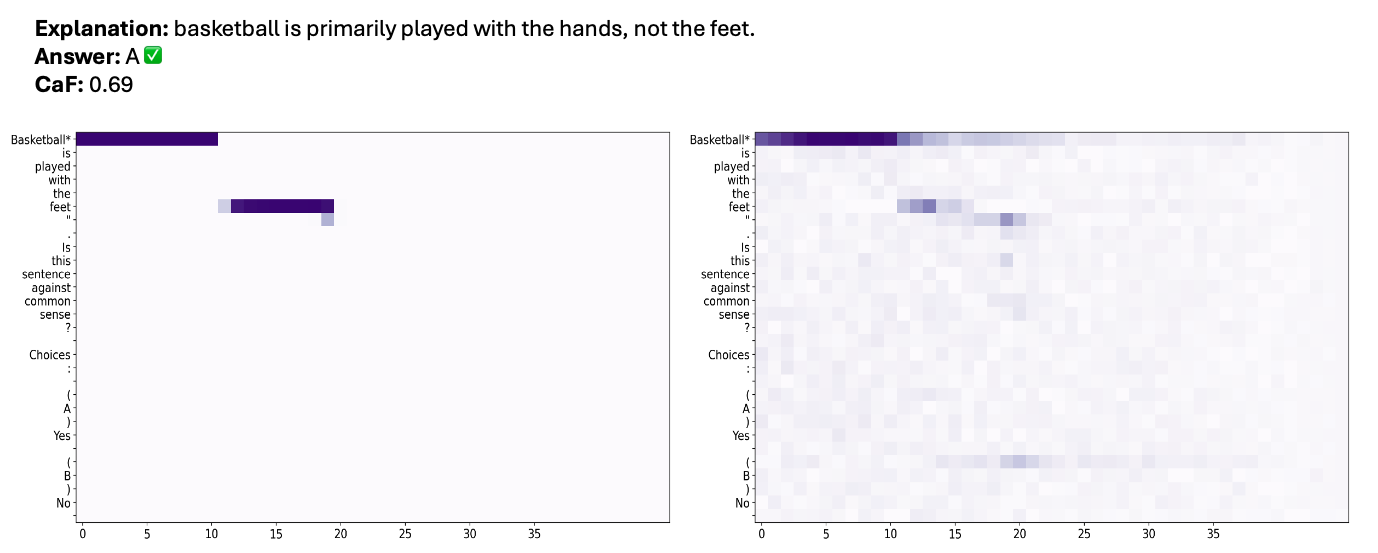}
        \caption{Gemma2-27B-chat}
    \end{subfigure}
    \caption{ComVE: Gemma2-27B LLMs.}
    \label{fig:esnli_27b}
\end{figure*}

\end{document}